\documentclass{article}

 \usepackage[main, final]{neurips_2026}

\usepackage[utf8]{inputenc} 
\usepackage[T1]{fontenc}    
\usepackage{hyperref}       
\usepackage{url}            
\usepackage{booktabs}       
\usepackage{amsfonts}       
\usepackage{nicefrac}       
\usepackage{microtype}      
\usepackage{xcolor}         

\usepackage{graphicx} 
\usepackage{tabularx}
\usepackage{xltabular}
\usepackage{subcaption}
\usepackage{siunitx}
\usepackage{csquotes}
\usepackage{xspace}
\usepackage{enumitem}

\usepackage{stfloats}

\usepackage{rotating}  
\usepackage{adjustbox}
\usepackage{booktabs}
\usepackage{multirow}

\newcommand{\omt}[1]{}

\providecolor{blue_plain}{rgb}{0,0,1}
\providecolor{red_plain}{rgb}{1,0,0}
\newcommand{\unc}[1]{{\color{blue_plain}{}#1}}
\renewcommand{\unc}[1]{\relax}

\newcommand{\todo}[1]
{{\color{blue_plain}{}TODO: #1}}

\renewcommand{\todo}[1]{}

\newcommand{\cbrk}{\protect\\}

\definecolor{bg_gray}{RGB}{245,245,245} 
\definecolor{frame_gray}{RGB}{220,220,220} 

\usepackage{float}
\usepackage[most]{tcolorbox}
\usepackage[table]{xcolor}
\newfloat{prompt}{htbp}{lop}
\floatname{prompt}{Prompt}
\definecolor{googleblue}{RGB}{66, 133, 244}
\definecolor{googlegray}{RGB}{245, 245, 245}
\definecolor{darkgray}{RGB}{60, 60, 60}

\newtcolorbox{promptbox}[1][]{
  enhanced,
  colback=white,
  colframe=googlegray!80!black, 
  arc=3pt,
  boxrule=0.5pt,
  left=5pt, right=5pt, top=5pt, bottom=5pt,
  fontupper=\small\sffamily, 
  title=#1,
  coltitle=black,
  colbacktitle=googlegray,
  attach boxed title to top left={yshift=-3mm, xshift=3mm},
  boxed title style={boxrule=0pt, colframe=white, colback=googlegray},
}

\usepackage{listings}
\newtcbox{\var}[1][]{on line,
  arc=2pt, colback=googleblue!10!white, colframe=googleblue!50!white,
  boxrule=0.5pt, boxsep=0pt, left=2pt, right=2pt, top=1pt, bottom=1pt,
  fontupper=\small\ttfamily\color{googleblue!60!black},
  #1
}
\definecolor{choiceorange}{RGB}{230, 120, 0}
\newtcbox{\choice}[1][]{on line,
  arc=2pt, colback=choiceorange!10!white, colframe=choiceorange!50!white,
  boxrule=0.5pt, boxsep=0pt, left=2pt, right=2pt, top=1pt, bottom=1pt,
  fontupper=\small\ttfamily\color{choiceorange!60!black},
  #1
}
\usepackage{fontawesome5}
\newcommand{\user}{%
  \vspace{0.5em} 
  \noindent\textbf{\color{googleblue} \faUser\ User:} 
}

\newcommand{\model}{%
  \vspace{0.5em}
  \noindent\textbf{\color{darkgray} \faRobot\ Model:} 
}

\title{Bouba Beats Kiki, and Other Regularities in the Subjective Preferences of LLMs}

\author{
Katy Blumer \\
Cornell University / Google\\
\texttt{kblumer@cs.cornell.edu} \\
\And
Jon Kleinberg \\
Google \\
\texttt{kleinberg@cornell.edu} \\
\AND
Ravi Kumar \\
Google \\
\texttt{ravi.k53@gmail.com} \\
\And
Andrew Tomkins \\
Google \\
\texttt{atomkins@gmail.com} \\
}
\begin{document}
\addtocontents{toc}{\protect\setcounter{tocdepth}{-1}}

\maketitle

\begin{abstract}
Users engage with large language models on tasks that are purely objective (``When was Netflix founded?''), purely subjective (``What should I name my child?'') or anywhere in-between. We have rich tools for evaluating model accuracy with respect to objective criteria. However, when it comes to subjective aspects of a task, we do not have similarly developed methodology at hand. In this paper, we take some early steps towards framing the question of subjective evaluation. Rather than measure how often a model is right or wrong on a question, we instead ask, how do models make choices among multiple viable responses without a clear winner?

To study this question, we restrict our focus to short-form content, and consider comparison tasks among three such content types: numbers, baby names, and fictional cities. We identify 29 aspects of these short-form examples covering many areas of syntax and semantics (length, alphabetic order, popularity, vowel mix, spoken sound, etc.). We show first that models tend to have strong and reproducible preferences for or against some of these aspects. Second, we show that variations in how the model is asked to express a preference may cause some of these aspects to activate or deactivate in the model's selection. Third, we show that overall preference distributions among pairs may vary radically based on seemingly-unimportant variations in the nature of the question posed to the model, raising new questions about manipulation. And finally, we show that, if models are used both to create lists of alternatives and to rank them, there are unexpected interactions between the two stages that may also drive the models' stated preferences.
\end{abstract}

\section{Introduction} \label{sec:intro}


LLMs have demonstrated remarkable advances in domains that admit clean signals of correctness to guide training. These signals come from a range of sources, including:
\begin{itemize}[nosep]
    \item Large corpora used in early training for next-token prediction, e.g.,~\citep{dolma2024}.
    \item Human rating data or expert-created content used for SFT or RLHF~\citep{ouyang2022}.
    \item Model-derived  auto-ratings used for reward modeling or RLAIF~\citep{ouyang2022}
    \item Automated output validation as in coding tasks with complete tests~\citep{swebench2024} or math problems with Lean proofs~\citep{lean2021}, or logic puzzles with known correct answers, used  in training as RLVR~\citep{rlvr2025,deepseek2025}.
\end{itemize}

Under this pattern, models are optimized to produce outputs that are ``preferred'' under various criteria like those described above. This highly successful recipe works because it relies on a purely objective comparison of task outputs. In reality, the tasks that humans give to LLMs contain both objective and subjective factors. Our focus in this paper is to explore the subjective elements that do not have clear winners and losers, but are nonetheless critical to model performance. Consider for example the task of naming a newborn child. Expecting parents view this task as important and increasingly ask LLMs for input, but we do not approach this task by producing an objective order of baby name quality; instead, we recognize that there are critical subjective factors present.

In domains such as innovating business plans~\citep{DeFreitas2025} or research directions~\citep{Lu2025,Si2024}, domain experts may be invoked to evaluate the quality of model outputs, but this expert evaluation does not make the task purely objective in nature. First, expert feedback is expensive and may not available for all tasks at the volumes required for training. Second, the feedback itself may be subject to decisions made by the researchers about factors whose relative importance are unknown: how important is novelty compared to capital requirements, for instance?

To understand LLM behavior with respect to the subjective elements of a task, we are inherently looking beyond the objective matter of which output is better. We may still evaluate the contribution of the LLM, but rather than asking whether the LLM found the ``right'' answer, we ask instead what conditions influence the model's choice among alternatives that are acceptable but entirely comparable. If certain models, or approaches to invoking those models, may produce radically different outcomes, we wish to understand the nature of these differences. Does one invocation more heavily emphasize particular elements of the problem? Do some invocations introduce more randomness than others? Are model choices disproportionately impacted by presentation? How do models respond to prompt changes that humans view as immaterial? What other factors are at play?

To do this study we must choose where to focus in the broad world of tasks with subjective elements. An important distinction can be drawn between short-form tasks, such as the baby name choice describe above, versus longer-form tasks, like crafting a social media bio. We see long-form tasks as inherently more difficult to study, as there are subtle interactions between the words in the generated output that make it difficult to unearth how specifically the model chose to distinguish between two long-form alternative. To keep the problem framing as simple as possible, we focus our attention on a range of short-form tasks, and find that, even in this setting, there is a wealth of complexity to study. We experiment specifically with on baby names, the names of fictional cities, and choices of numbers.



\smallskip\noindent\textbf{Problem and methodology.}
We propose methods for investigating
the factors that impact the LLM's decision in subjective settings without any clear right answer.
There is a mature academic literature characterizing the traditional ``cognitive biases'' that can arise when humans undertake subjective tasks. It is unclear, however, what corresponding forms of ``LLM reasoning biases'' might similarly influence model outcomes in subjective brainstorming.\footnote{We will use ``bias'' in the statistical sense going forward to indicate model preferences for certain classes of outputs.  We do not mean to imply that biases represent reasoning errors; instead, they may be viewed as dimensions the model may focus on in making a choice. We understand this might suggest an unintended connection to \emph{model bias}, but we have not been able to identify an alternative term for the concept.}
In all, we study comparison tasks in which a model is asked to select an alternative from two choices. We identify and study a total of 29 biases that may exist in these choices, across several domains, plus an additional set that emerges when the model is also used to create the list of items from which the comparison pairs are drawn. As examples of the types of biases we consider, a model might prefer names that are longer or shorter, that begin with hard consonant sounds, or that are evocative of professions associated with agrarian societies. 


To make our insights as generalizable as possible
we design the domains and the tasks to be structured and stylized.
We use three distinct structured domains representing different types of short-form comparable objects: human names (pulled from the most popular baby names in the US Census), fictional city names (generated by an LLM), and numbers (randomly sampled as a baseline).
For city names, we set up the subjective decision-making framework through two interrelated tasks:
\begin{itemize}[nosep]
\item \textsl{List creation}: given a target output (such as a fictional city name or a product tagline),
the LLM should produce a list of $N$ options for this target output.
\item \textsl{Comparison}: given two options from the list, the LLM should say which option it favors.
\end{itemize}
Our comparison task relates to past work on ideation~\citep{li2025ideation} and on biases
in LLM-as-a-judge frameworks~\citep{bai2023benchmarking,li2025generation,zheng2024judging},
and we consider these further in Appendix~\ref{app_sec:related_work}.
But we extend the analysis in several new directions, as we discuss next.

\smallskip\noindent\textbf{Analysis.}
First, we analyze how the multiple LLM biases we have identified interact
in driving outcomes for this basic comparison task. We include well-studied biases based on
position (which option is presented first and which is presented
second)~\citep{shi2024judging} 
and alphabetical ordering, along with many others. We also consider biases that depend on the execution of the list creation phase: consider for example that a model might be biased in favor of options generated earlier in the list, as these items might be more readily accessible to the model. As this example suggests, list creation and comparison may interact with each other even when formulated as distinct tasks.

We also consider the prompt used to drive the model's comparison of two items. We find that relatively small changes to the prompt---even when the instructions appear essentially identical---can induce non-trivial changes in the outcomes. In particular, we observe that the choice of prompt can excite or depress the influence of certain biases in the LLM's comparison outputs. Sometimes a careful reading of the prompt can explain these biases. For example, asking \enquote{Which item would you select first?} instead of \enquote{Which item do you prefer?} might activate the bias in favor of alphabetically earlier items. More often, though, we do not discern any obvious reason why a particular prompt would cause one bias or another to become a stronger driver of the model's outcome.
The fact that innocuous changes to the prompt might have a significant influence on outcomes raises both new research questions and new questions about LLM behavior.
%

Finally, we identify another source of interaction between list creation and comparison.
In list creation, we find that the LLM will often generate options that fall
into natural categories. In generating names
of cities, e.g., there may be a run of contiguous city names in the list
that are based on seasons, another contiguous run of city names
that are based on compass directions, and so forth.
We then find that many prompts will induce biases that favor
members of certain categories over others; this therefore is a bias
that cannot be expressed in terms of fixed a priori features of the
options, but instead in terms of the implicit categories they belong
to. Hence, understanding model comparisons among choices also requires understanding the hidden structure behind the creation of the candidates.

Taken together, our results provide a richer view of some of the
empirical regularities in how LLMs engage in subjective
tasks, both in generating option lists and in comparing items from those lists.

\section{Methods}
\label{sec:methods}

\smallskip\noindent\textbf{Comparable objects.}
As discussed in Section~\ref{sec:intro}, we work with short-form items so that the features are more clear-cut, and less dependent on more complex textual or stylistic properties associated with longer passages of text.
Three types of short-form items were used: numbers, human names, and fictional city names.
Numbers were generated by sampling 200 unique numbers from $\{0, \dots, 999\}$.

Human names (henceforth referred to just as ``names'') were generated by pulling the most common 100 female and male names from the 2020 US Census~\citep{Comenetz2026}. Names were defined as ``female'' or ``male'' based on the ``counts by sex'' columns in the provided data; all names in the most common 100 of each category had at least a 60/40 gender split.

Fictional city names (henceforth referred to as ``cities'') were generated by asking Gemini 2.5 Flash for a list of 200 fictional city names. A duplicate was later removed, yielding 199 cities; see Appendix~\ref{app_sec:comp_object_tables}

\smallskip\noindent\textbf{Prompt expansion.}
The main set of 50 prompts was generated using iterative expansion from one prompt. The seed prompt was \enquote{Which of these two names do you prefer?.} Then an LLM was asked to paraphrase the seed prompt 10 times (Prompt~\ref{prompt:prompt_expander}). This process was repeated 5 times, each time giving the LLM the set of already-generated prompts to avoid duplicates. The same 50 prompts were used for all 3 datasets, replacing  \enquote{names} with \enquote{numbers} or \enquote{cities} as appropriate; see Appendix \ref{app_sec:prompt_tables}.

\smallskip\noindent\textbf{Combination.}
Unless otherwise specified, results come from a dataset created as follows:
\begin{enumerate}[nosep]
\item 200 comparable objects from a given domain (numbers, names, etc.) are acquired or generated as described above.
\item A set of 200 unordered pairs is generated such that each object is paired with exactly two other objects. This is done by sampling a random cycle over the 200 objects. Specifically, we generate a set of 200 unordered pairs $\{obj_i, obj_j\}$ by randomly permuting the objects into a sequence and pairing adjacent elements---including an edge between the last and first objects---such that each object appears in exactly two pairs.
\item Each unordered pair $\{a, b\}$ is compared twice, once for ordering $(a, b)$ and for $(b, a)$.
\item Each ordered pair is compared using each of 50 similar questions (\enquote{prompts}), structured using the standardized comparison template detailed in Appendix~\ref{app_sec:comparison_prompt_template}.
\end{enumerate}

This produces a total of $200 \times 2 \times 50 = \num[group-separator={,}]{20000}$ LLM-driven comparisons.

Comparisons were all done with a temperature of 1 and model defaults for thinking, unless otherwise specified; see Section~\ref{app_sec:prompt_tables}. The set of LLMs that were queried is in Table~\ref{tab:A.1_bias_fracs}; details of their exact configurations, API identifiers, and citations are in Appendix~\ref{app_sec:model_configs}.
For reasons of space, many figures show only results for Gemini 2.5 Flash, but results for other SOTA models are broadly similar and are in Appendix~\ref{app_sec:other_models_figs} (prompt histograms in Appendix~\ref{app_sec:prompt_hist_others} and across-model correlations in Appendix~\ref{app_sec:prompt_corr_models}).

\section{Experimental Results} \label{sec:results}

We begin in Section~\ref{sec:bias} by taking the features we have identified and studying how much each of them contributes to the overall  comparison outcomes produced by the LLM. 
Next in Section~\ref{sec:prompts} we slice these overall influence values by prompt, to see how much the wording of the prompt changes the outcomes. Finally, in Section~\ref{sec:ideation} we revisit the ideation phase and discuss how domain-specific categories, used by the LLM to generate items, go on to influence the outcomes of choices.

\subsection{Observed Preferences}\label{sec:bias}

We set about to develop a robust but non-comprehensive set of features that would allow us to study LLM behavior in comparisons. Through an iterative process that began with known features, and expanded the set using iteration and suggestions proposed by Gemini 2.5 Flash in response to prompting, we identified a few dozen exogenous features, some of which are specific to a single domain, while others apply across two or three domains. 
The features for the names and numbers datasets are listed in
Table~\ref{tab:A.1_bias_fracs} (brief definitions for unfamiliar terms are provided in Appendix~\ref{app:bias-list}). Metrics for cities are listed in Appendix Table~\ref{tab:A.3_bias_fracs_cities}. 
As the asterisks indicate (using a Bonferroni correction for multiple comparisons across all model-metric pairs, $m=183$), many features occur on the winning item in a comparison significantly more or less often than random chance would suggest. We evaluated statistical significance against a 0.5 baseline using a one-sample t-test over the win rates of the $N$ unique unordered pairs.
 Crucially, these reported $p$-values are conservative estimates. To avoid inflating significance through correlated outcomes, we define our sample size $N$ strictly as the number of unique unordered pairs. Consequently, each pair contributes only once to the statistical test, despite representing up to 100 individual comparisons (2 orderings $\times$ 50 prompts) in the dataset.

To assess the contribution of each feature to the observed outcomes, we report two measures for each feature. First, we compute for each feature $i$ the probability that an item with feature $i$ wins a comparison it participates in against an item $j$ where the feature favors $i$ over $j$; this is the measure reported in Table~\ref{tab:A.1_bias_fracs}. However, as there are strong correlations among the features, we wish to tease out whether the feature $i$ actually contributes to the win, or simply appears alongside other driving preferences.
For this, we fit a multinomial logistic regression whose features are normalized scalar variables (length, order of magnitude) or binary variables indicating whether an item contains each preference (presented first, starts with a vowel, etc), so the item's odds of winning a comparison is proportional to $e^{w\cdot \beta}$ for binary feature vector $w$ and learned coefficient vector $\beta$. This is equivalent to assuming that each item's odds of winning a comparison are a product of odds multipliers $e^{\beta_i}$ for each feature $i$ present for the item. For ease of presentation, we report $e^{\beta_i}/(1+e^{\beta_i})-0.5$, the likelihood that an item with feature $i$ will win a comparison against an equivalent item without feature $i$, so it is directly comparable to the standalone measure we described above. 

A causal decomposition using minimal pairs to validate these coefficients is in Appendix~\ref{app_sec:minimal_pairs}. To distinguish prompt-level variance from sampling noise, we also present a mixed-effects analysis in Appendix~\ref{app_sec:mixed_effects}.

\begin{table}[h]
\centering
\caption{Summary of model preferences across different features for the ``Names'' and ``Numbers'' domains. ``Preference" here means the probability that an item $i$ wins a comparison against an item $j$, given that the feature favors $i$ over $j$.
\cbrk
``Random Pref Order'' represents a simple baseline (for the full Monte Carlo version of this baseline, see Appendix~\ref{app_sec:monte_carlo}). For each of these 3 columns, we assign a random preference ordering over all items in the domain, to simulate some unknown ``true'' preference ordering that could be driving the observed preferences. We then decide all matchups according to that ordering, and then compute win rates for each feature. 
\cbrk
Ties (cases where the feature favors neither item) are excluded. Features where over half of the matchups were excluded are italicized, with the percentage of active (non-excluded) matchups listed in parentheses after the feature name. Significance stars reflect Bonferroni-corrected $p$-values across $m=183$ cell tests ($^{***} p_{\text{Bonf}} < 0.001$, $^{**} p_{\text{Bonf}} < 0.01$, $^{*} p_{\text{Bonf}} < 0.05$).}
\label{tab:A.1_bias_fracs}
\resizebox{\textwidth}{!}{
\begin{tabular}{lllllll||lll}
\toprule
 &  & \shortstack{Gemini \\ 2.5 Flash} & \shortstack{Gemini \\ 3 Flash} & \shortstack{Claude \\ Haiku 4.5} & \shortstack{GPT-5 \\ Mini} & \shortstack{Qwen \\ 3.5 35B} & \shortstack{Random \\ Pref Order 1} & \shortstack{Random \\ Pref Order 2} & \shortstack{Random \\ Pref Order 3} \\
Comparable & Metric &  &  &  &  &  &  &  &  \\
\midrule
\multirow[c]{12}{*}{Numbers} & Higher & {\cellcolor{white}} \color{black} {\cellcolor[HTML]{3B4CC0}} \color[HTML]{F1F1F1} 0.70\textsuperscript{***} & {\cellcolor{white}} \color{black} {\cellcolor[HTML]{3B4CC0}} \color[HTML]{F1F1F1} 0.78\textsuperscript{***} & {\cellcolor{white}} \color{black} {\cellcolor[HTML]{3B4CC0}} \color[HTML]{F1F1F1} 0.79\textsuperscript{***} & {\cellcolor{white}} \color{black} {\cellcolor[HTML]{3B4CC0}} \color[HTML]{F1F1F1} 0.76\textsuperscript{***} & {\cellcolor{white}} \color{black} {\cellcolor[HTML]{3B4CC0}} \color[HTML]{F1F1F1} 0.72\textsuperscript{***} & {\cellcolor{white}} \color{black} {\cellcolor[HTML]{F2CBB7}} \color[HTML]{000000} 0.48 & {\cellcolor{white}} \color{black} {\cellcolor[HTML]{8CAFFE}} \color[HTML]{000000} 0.55 & {\cellcolor{white}} \color{black} {\cellcolor[HTML]{C53334}} \color[HTML]{F1F1F1} 0.41\textsuperscript{*} \\
\bottomrule
\end{tabular}

}
\end{table}


\begin{figure*}[htb]
    \centering
    \includegraphics[width=1\linewidth]{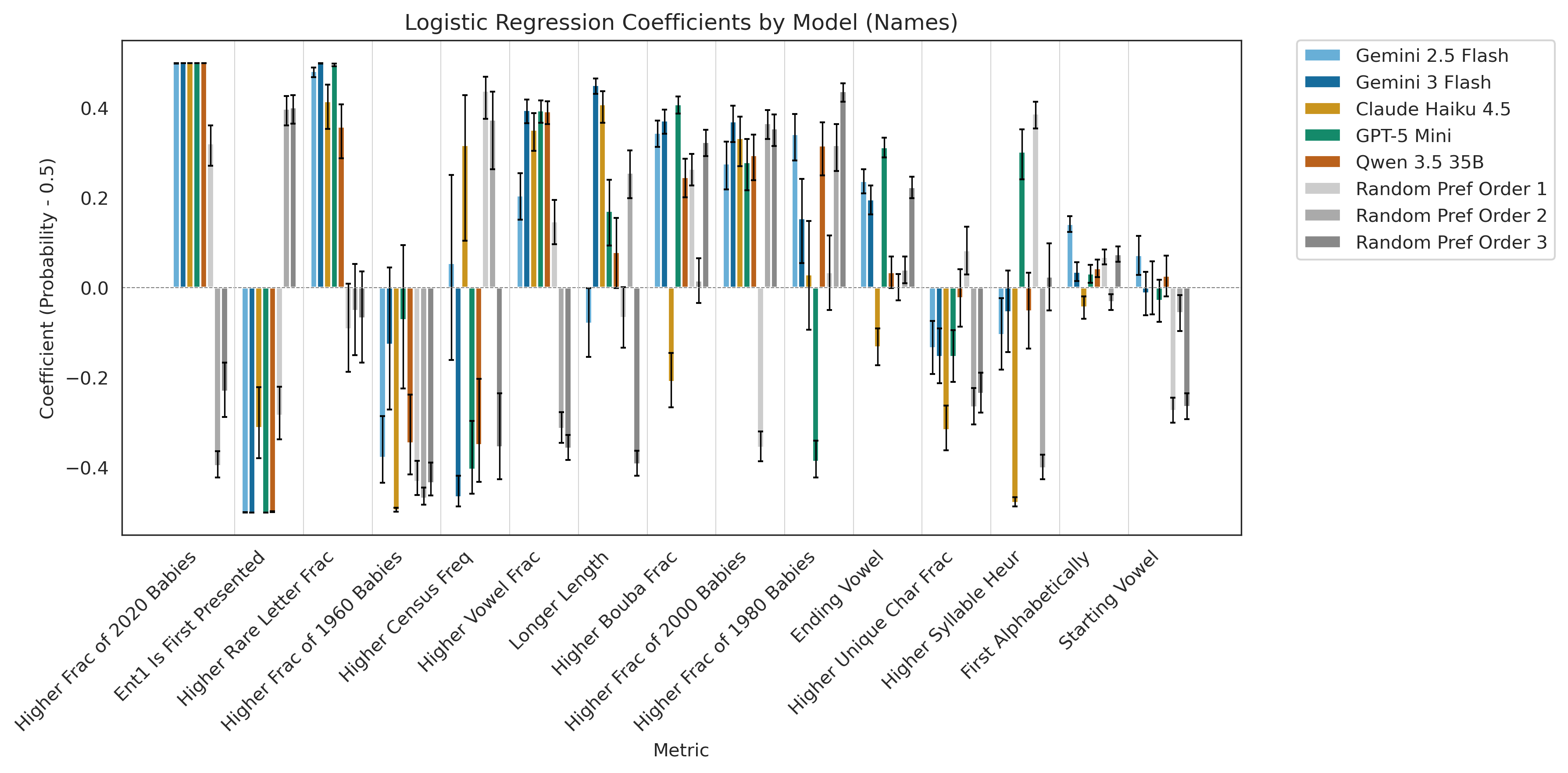} \caption{Second preference measure: logistic regression coefficients (with 95\% confidence intervals), predicting whether an item wins a comparison against another. Presented along with overall win fractions from Table~\ref{tab:A.1_bias_fracs},  both scaled to the $(-0.5, 0.5)$ range. Positive values mean a feature is positively correlated with likelihood of winning.}
\label{fig:B_log_reg}
\end{figure*}

\smallskip\noindent\textbf{Findings.}
Results for names are shown in Figure~\ref{fig:B_log_reg} (results for numbers and cities in Appendix~\ref{app_sec:regression_coeffs_others}). Notice first that the preference values in Table~\ref{tab:A.1_bias_fracs} are sometimes larger and sometimes smaller than the logistic regression coefficients. This is to be expected, as the likelihood of winning when a feature is present may regress to the mean due to the noise of other preferences; on the other hand, the logistic regression coefficients may similarly regress towards 0.5 because the model may choose to move weight to more predictive but highly correlated other features. Nonetheless, some patterns emerge:

Across the evaluated models and datasets, several preferences exhibit highly consistent behaviors, while others show stark divergences depending on the domain or specific model family. Among consistent trends, models show a strong and statistically significant preference for larger numbers (win rates of $0.70$--$0.79$) and prime numbers ($0.64$--$0.73$) in the numbers domain, as well as names popular in recent decades (e.g., 2020 Babies) over historical ones (1960 Babies, win rates $<0.42$). Phonetically, a shared preference emerges across both names and fictional cities for starting vowels (up to $0.76$ for names, $0.64$ for most cities), higher vowel fractions, and soft ``bouba'' sounds ($>0.57$).

Conversely, other features reveal sharp divergences between domains or models. While models are indifferent to the length of human names, they exhibit an aversion to longer fictional city names (win rates $<0.43$). In numbers, while most models are indifferent to divisibility, Claude Haiku 4.5 and GPT-5 Mini significantly avoid multiples of 10 ($0.37$), 5 ($0.40$), and even numbers ($0.41$). Presentation order effects (``first presented'') also diverge significantly: names exhibit a consistent primacy preference across all models ($0.82$ for Haiku), while cities show a strong recency preference for Gemini 3 Flash ($0.47$) and GPT-5 Mini ($0.39$). Finally, list position in cities reveals a hierarchical contrast, where models generally prefer later-generated categories overall (win rates $<0.44$ for Gemini models) but prefer earlier items within those specific categories ($0.56$ for Gemini 3 Flash).

\subsection{Prompt variation} \label{sec:prompts}
We move now to a study of the difference in outcomes as we move from one prompt to another. For each of our three domains, we pick a set of important features. Then over the fifty prompts, we compute separately for each prompt the \emph{winning fraction}; that is, fraction of comparisons won by instances having the feature against items without the feature. We then show the distribution of winning fraction. A highly concentrated distribution means that all prompts behave similarly with respect to this preference. A highly dispersed distribution means that some prompts cause the model to prefer items with this feature, while other prompts prefer items \emph{without} this feature. This computation therefore gives us a sense of the amount of influence a prompt can have in activating or supressing the particular preference.

\begin{figure*}
    \centering
    \includegraphics[width=1\linewidth]{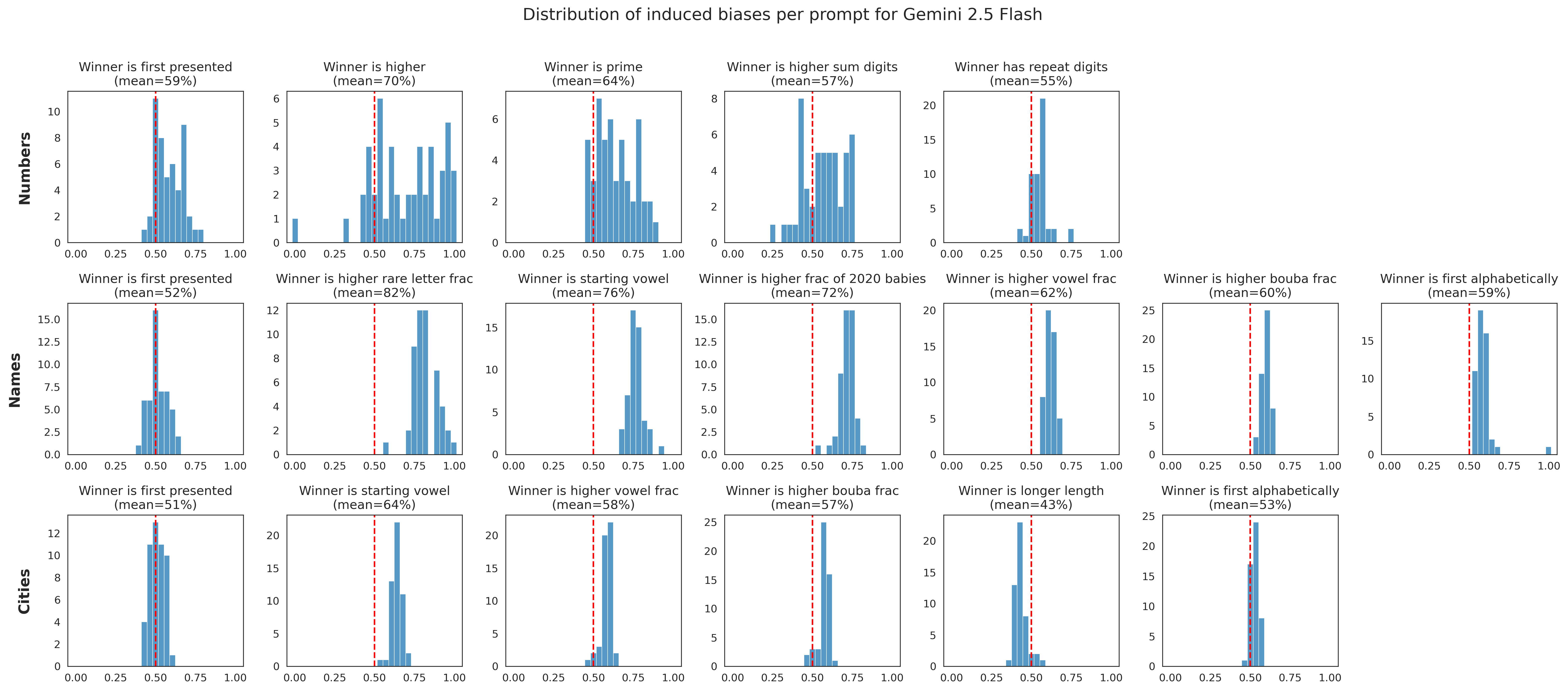}
    \caption{Prompt variation: distribution of per-prompt means for each preference (probability item $i$ wins a comparison against an item $j$ where the feature favors $i$ over $j$). }
    \label{fig:D_prompt_bias_hist}
\end{figure*}

The results are shown in Figure~\ref{fig:D_prompt_bias_hist}. First, many of the graphs seem to have a bell curve shape. This is natural, as each prompt's winning fraction is the result of $200*2$ random trials, so there will be some expected dispersion. The first question is whether the dispersion of the winning fraction can be explained by this randomness, even under the null hypothesis that the prompts themselves do not operate to strengthen or weaken a preference.
To evaluate this question, we computed Levene's statistic to ask whether the winning fraction distributions are consistent with the expected natural noise. In all the twenty-one distributions in the figure, this null hypothesis is refuted with $p<0.001$ (and typically far smaller). So we conclude that the choice of prompt does indeed control the activation or suppression of each preference.

The dispersion in the distribution of winning fractions for numbers is significantly higher for most preferences than for names and cities. We give some more details on this observation below, but we may conclude that particular domains may be more or less susceptible to prompt influence, though all three domains we study show significant influence.

\begin{figure*}
    \centering
    \begin{subfigure}{0.48\linewidth}
      \includegraphics[width=1\linewidth]{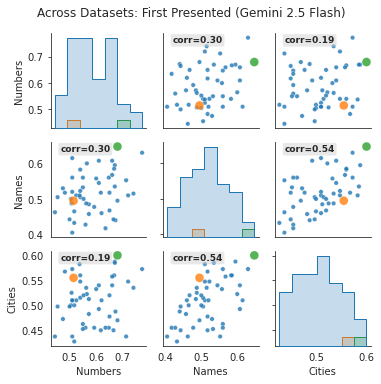}
      \caption{First Presented}
    \end{subfigure}
    \hfill
    \begin{subfigure}{0.48\linewidth}
      \includegraphics[width=1\linewidth]{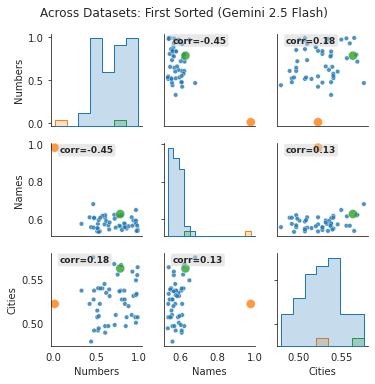}
      \caption{First Sorted}
    \end{subfigure}
    \caption{Robustness of prompt variation: correlation of prompt preference between datasets on two different metrics. \enquote{First sorted} refers to first alphabetical for cities and names, and higher number for numbers. Prompts 23 and 18 are highlighted in orange and green respectively as illustrative outliers (see text for detailed discussion).}
    \label{fig:E_robust}
\end{figure*}

Prompts also show some correlation between datasets, as seen in Figure~\ref{fig:E_robust}; if an individual prompt shows a preference towards the first-presented option for names, for example, it is likely to show the same preference for cities. Similar cross-dataset correlations are observed for Gemini 3 Flash and GPT-5 Mini, as detailed in Appendix~\ref{app_sec:prompt_robustness_others}.

Some prompt preferences are more interpretable than others. For example, the most extreme outlier visible on the \enquote{first sorted} plot in Fig.~\ref{fig:E_robust} is prompt 23, shown in orange: \enquote{Regarding these two {cities/names/numbers}, which one would you put first?} It's perhaps not surprising that the model overwhelmingly answers that question with the option that comes first alphabetically, in the case of names or cities; or first on a number line, in the case of numbers. But another occasional outlier, prompt 18 (in green), is less clear: \enquote{Please identify your preferred option among these two {cities/names/numbers}.}

Different prompts also show differences in which factors are most salient in the model's thinking, as identified by feeding each comparison's thinking trace to a second LLM (Appendix~\ref{app_sec:thinking_names}).

\begin{figure*}[htb]
    \centering
\begin{subfigure}{0.7\linewidth}
    \centering
    \includegraphics[width=1\linewidth]{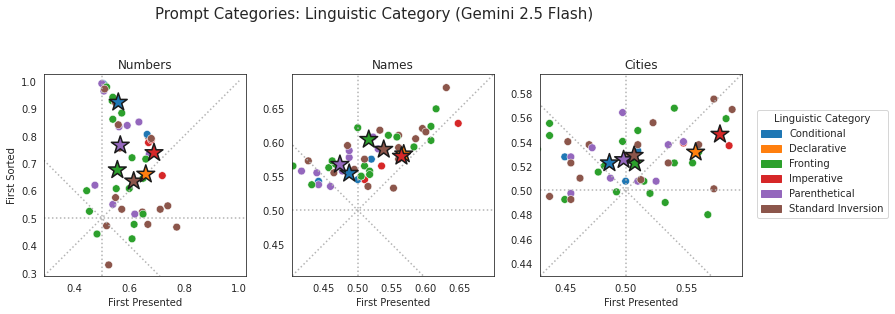}
    \caption{Prompt preferences by \enquote{linguistic category}, as defined by an LLM (see Table~\ref{tab:prompt_cat_grammatical_explanation} for definitions).}
    \label{fig:G}
\end{subfigure}
\begin{subfigure}{0.7\linewidth}
    \centering
    \includegraphics[width=1\linewidth]{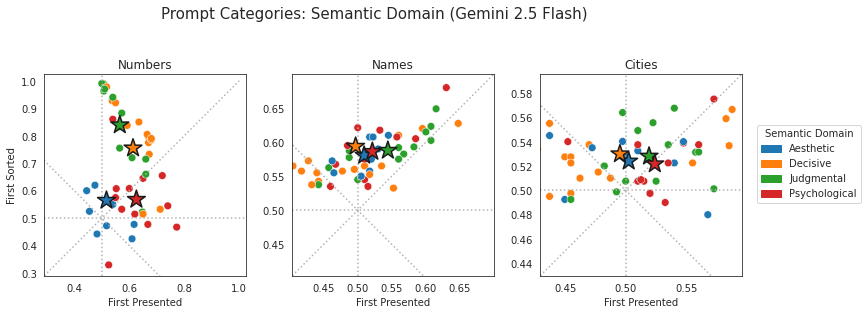}
    \caption{Prompt preferences by \enquote{semantic domain} of the prompt's verbs and nouns, as defined by an LLM (see Table~\ref{tab:prompt_cat_semantic_domain} for definitions).}
    \label{fig:H}
\end{subfigure}
\caption{\enquote{First-sorted} (alphabetical or numerical) vs.\enquote{first-presented} preferences per prompt, showing prompts colored by various categorizations. The centroid of each category is shown as a star.}
\label{fig:GH}
\end{figure*}

Some prompt preferences may stem from more subtle features. For example, Figure~\ref{fig:G} shows that prompts with "imperative" structure seem to have a stronger preference towards the first-presented option, while sentences with parentheticals may bias the model more towards the second-presented option, at least for words.
 Figure~\ref{fig:H} shows that prompts using words judged to belong to the \enquote{Decisive} or \enquote{Judgmental} domains show stronger preferences than questions deemed \enquote{Aesthetic} or \enquote{Psychological}. (Categories were determined by a second LLM which was given only the raw list of 50 prompts. See Appendix \ref{app_sec:prompt_tables}
 for category definitions and a table of prompts.)

\subsection{Ideation} \label{sec:ideation}
We note that while these endogenous category structures are particularly pronounced in lists generated by Gemini 2.5 Flash, they represent a common property of how LLMs serialize long lists of subjective candidates. Newer SOTA model families (such as Gemini 3 or GPT-5) often employ strong implicit structural priors—sometimes explicitly partitioning a requested list of 50 items into structured sub-categories (e.g., 5 themed categories of 10 items each) even when not requested. In this sense, the category-based "spillover" preference we observe is not merely an artifact of an older model, but rather suggests that structural serialization preferences persist, and in some cases become more formalized, in newer architectures.
\begin{figure*}
    \centering
    \includegraphics[width=0.9\linewidth]{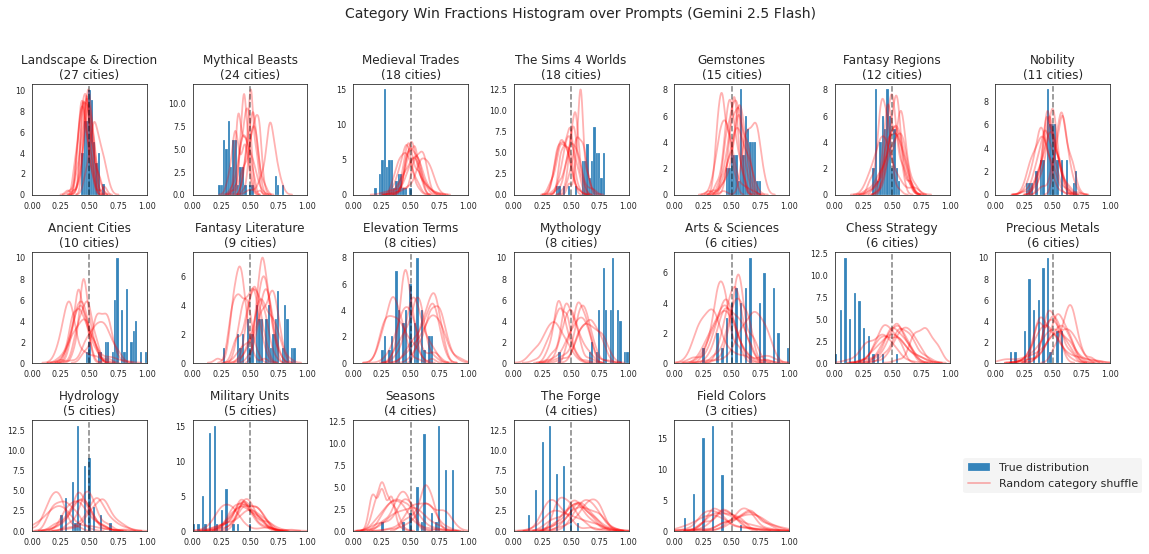}
    \caption{Category preference: distributions over per-prompt win probabilities for each category (probability that a city from a given category wins against a city from any other category). Plotted against baseline (red lines) of the same distribution, calculated for 10 different random assignments of cities to categories. See Table \ref{tab:J_city_cats} for a list of city names and categories.}
    \label{fig:K}
\end{figure*}


Recall that the stylized {\em ideation} activity we define is to produce an initial list of candidate options, as described in Section \ref{sec:methods}. The candidates produced by this list-creation task are then the objects that get compared in the comparison task that we have been studying in this section.

These two tasks are related, not just by the fact that the list-creation task produces the objects for the comparison task, but by the fact that more fine-grained properties of the list-creation task spill over into the outcome of the comparison task. We have seen a first example of this in the way that the ``first-generated'' attribute is a source of preference: the model has a preference for fictional city names that were generated earlier in the list of candidates rather than later. Here we discuss a more subtle form of spillover, in which the nature of the list itself can create a set of {\em endogenous} categories that introduce new forms of preference.

This phenomenon is domain-dependent, but it shows up clearly with the Gemini 2.5 model and the domain of city names. Inspection of the list of 200 city names makes clear that contiguous cities have often been generated using a common theme; for example, cities 66 through 69 on the list are Springdale, Summerwind, Autumnwood, and Winterfell, suggesting that once the model embarked on this set of cities, it ran through the list of seasons. In fact, essentially the entire list of 200 cities can be broken into contiguous runs of thematically related objects, with the theme eventually pivoting to a new theme, and a new run beginning. Table~\ref{tab:J_city_cats} in the Appendix shows a Gemini-produced breakdown of the 200 cities into categories, in continguous runs of city names that were produced consecutively. Note that some of these categories (like city names beginning with seasons) are more obvious, whereas others require more specialized knowledge, even if they're equally clear-cut once Gemini surfaces this knowledge (like the run of 18 city names that come from The Sims 4~\citep{sims4wiki}).

These are the endogenous categories we referred to above, since they are only defined in terms of the actual choices of city names generated by the model; this is in contrast to attributes such as vowel fraction or syllable count which can be defined exogenously before the city names are seen. An additional set of preferences can be defined in terms of these endogenous categories: in the comparison task, we find that the model consistently prefers names from some endogenous categories over others. Figure~\ref{fig:K} details this category-based preference: for each category of city names identified by Gemini, the histogram shows how much each prompt shows a preference toward or against names from this category, based on win-rate against all other names. Relative to a random baseline (in which we randomly shuffle category memberships), certain categories are heavily favored across essentially all prompts (including ancient cities, mythology, and cities from The Sims 4), and certain categories are heavily disfavored across essentially all prompts (including city names based on medieval trades and chess strategy). Robustness checks confirming that these categories are content-driven rather than position-driven are provided in Appendix~\ref{app_sec:category_shuffling}.

In this way, we see that not only is there fine-grained structure in some of the ideation lists that are produced---in this case, the city names created by the model---but this fine-grained structure becomes a source of further preferences in the subsequent comparisons.


\section{Conclusion and Limitations}
\label{sec:conclusion}


In this paper we have studied two stylized tasks that capture aspects of the ideation and pruning phases in subjective decision-making: ideation inherently involves the creation of a list of options and pruning is built on the ability to decide whether option $A$ or $B$ is preferred. We study the preferences that arise as empirical regularities in this comparison process, and find that the strength and even the direction of a number of these preferences are prompt-dependent. 

And we identify some spillovers between the ideation and comparison tasks, in the following sense. When the model needs to create a long list of candidate options, it does so by iterating through high-level thematic categories and creating several members from each category. This in turn creates new preferences from the ideation phase, in which options from certain categories are preferred to options from other categories, again in a way that is mediated non-trivially by the prompt.

\paragraph{Limitations} 
As with all research in this area, it is useful to point out some of the limitations of our work. First, we chose a range of models that are at or near the state-of-the-art frontier, but our selection from within this set was oriented toward using smaller models where the cost of the computation would be lower. We sampled different styles of models, and used both models are more open and those that are more closed, including models from several of the largest providers of LLMs. A more extensive study could try a wider range of models, including some that are more powerful and expensive than what we have included here.

Similarly, we have studied a particular set of subjective comparison tasks designed to represent different aspects of the problem. The full set of subjective comparison tasks is vast, and subsequent work could easily explore further problems in this style. Our work here will be relevant to such future studies, in that it provides a road-map for the different categories of questions that can be asked in such settings, and set up in a way that can be adapted to other subjective comparison tasks.

\paragraph{Further directions} An interesting research direction is to extend the categorization of prompts further: we have found that structure variations in the prompt can lead to systematic shifts in the preferences, but we do not have a complete taxonomy of these effects. And it would also be interesting to extend our framework to more complex subjective tasks, such as the ideation and pruning of longer texts, where there is even more structure for the model to work with and potentially to use as the basis for its preferences. 

\clearpage

\bibliography{refs}
\bibliographystyle{plainnat}

\clearpage
\appendix


\section*{Table of Contents for the Supplemental Material}
\addtocontents{toc}{\protect\setcounter{tocdepth}{2}} 
\tableofcontents
\newpage


\section{Related Work} \label{app_sec:related_work}
One of the main literatures that our work connects to is
the line of research on LLMs as judges or evaluators.
A number of recent papers document specific statistical biases or ``preferences'', and one
of our goals in this work is to study the interplay of all these
preferences, and to identify new ones systematically, for the types
of subjective brainstorming tasks we consider here.
In particular, \citet{wang2024fair} demonstrate
the effects of positional bias, which is one of the effects
we study here,
and \citet{miles2023unfaithful} show
that chain-of-thought text produced by a model does not
necessarily correspond to the basis for its decision.  
For biases that involve other contexts
not directly related to our framework,
\citet{sharma2024sycophancy} show the effects of
sycophancy, in which agreement with the user is prioritized,
and \citet{koo2024benchmarking}
and \citet{Panickssery2024evaluators}
examine self-preference, in which a model has a tendency to
prefer its own outputs.
In more factual settings, again distinct from the subjective
questions we consider here,
\citet{wu2025style} show that LLMs can view grammatical or stylistic errors
as more severe than factual errors, and
\citet{shen2023evaluators} consider longer-form
tasks such as abstractive summarization.

Other lines of work relate LLM preferences to corresponding human
behavioral and cognitive biases.
\citet{hu2025bias} and
\citet{talboy2023challenging} study this in the context
of in-group bias, with
\citet{lei2024social} considering gendered language
and
\citet{gamboa2025social} considering
multilingual bias.
\citet{vankoevering2024random} study
the failure of even state-of-the-art LLMs to produce
pseudo-random sequences, instead exhibiting the types of biases
humans display when they try to generate random outputs.
\citet{jones2022cognitive} suggest
using this analogy with human behavioral and cognitive biases
as a research strategy for probing LLM behavior.

Motivated by the abundance of these biases, a parallel line of
work has proposed frameworks for measuring biases and for
evaluation in the presence of biases.
\citet{li2025generation} and \citet{bai2023benchmarking}
provide overviews of these activities, and
specific frameworks include Prometheus~\citep{kim2024prometheus},
PandaLM~\citep{wang2024pandalm}, and
Auto-J~\citep{li2023generative}.
These issues have also been considered in specific domains such as
assessing the quality of translations~\citep{kocmi2023translation}
and for the production of natural-language critiques~\citep{saunders2022selfcritiquing}.

For the ideation portion of our work, in which lists of candidate options are produced before being refined,
\citet{li2025ideation} provide a general survey of approaches to this problem.
There have been discussions of this process in a number of specific domains in both the academic literature and in general-audience writing, as suggested in the introduction, including
for business ideas~\citep{DeFreitas2025},
research ideas~\citep{Lu2025,Si2024},
baby names~\citep{brant2026couple,blumer202naming},
names for new companies and new products~\citep{allbusiness2023startup},
and names for characters when creating works of fiction~\citep{picklesimer2023characters}.
As noted at the outset, our work focuses on the shorter-form versions of the task, to be able to isolate properties of the options more concretely.

\newpage

\section{Supplemental Robustness Checks and Statistical Validation}
\label{app_sec:robustness_checks}

\subsection{Monte Carlo Permutation Test}
\label{app_sec:monte_carlo}

To test whether our results could be explained by some other, hidden factor that was the actual source of a model's preferences, we conducted a Monte Carlo permutation test. For each domain, we repeated the following procedure 25,000 times: we randomly shuffled the list of entities to simulate some hidden ``true'' preference ranking. We repeated the exact same pairwise comparisons, resolving each comparison according to which entity came first in the simulated preference ranking. Then we recorded the win rate for each feature based on the resolved comparisons. After repeating this 25,000 times, we had a distribution of win rates for each feature, representing the distribution of win rates you might expect for that feature if the model's true preference were based on something unrelated to that feature.

\begin{figure*}[htb]
    \centering
    \captionsetup[subfigure]{aboveskip=1pt,belowskip=0pt}
    \begin{subfigure}{0.77\linewidth}
      \centering
      \includegraphics[width=1\linewidth]{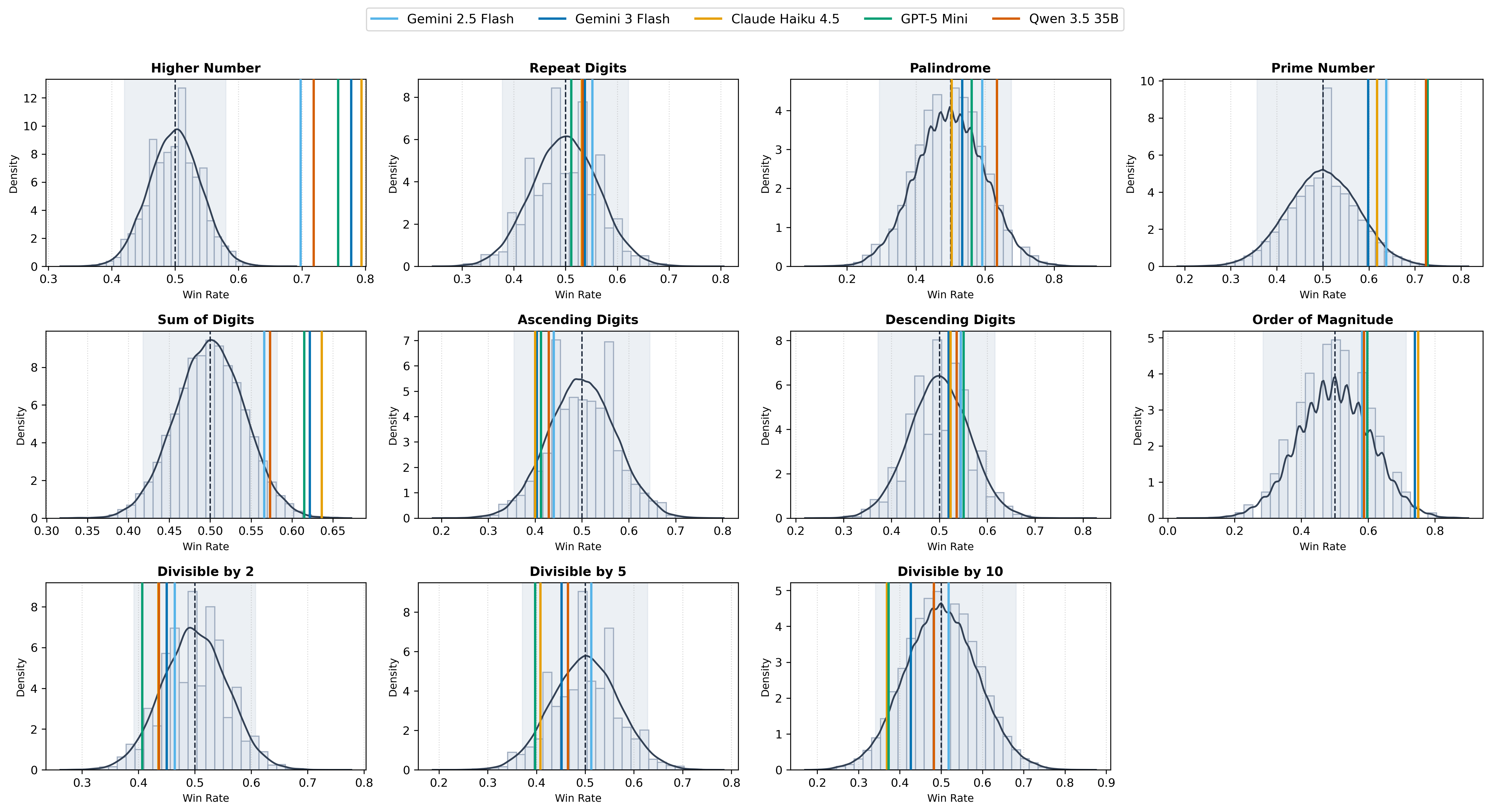}
      \caption{Numbers}
    \end{subfigure}
    \vspace{-4pt}
    \begin{subfigure}{0.77\linewidth}
      \centering
      \includegraphics[width=1\linewidth]{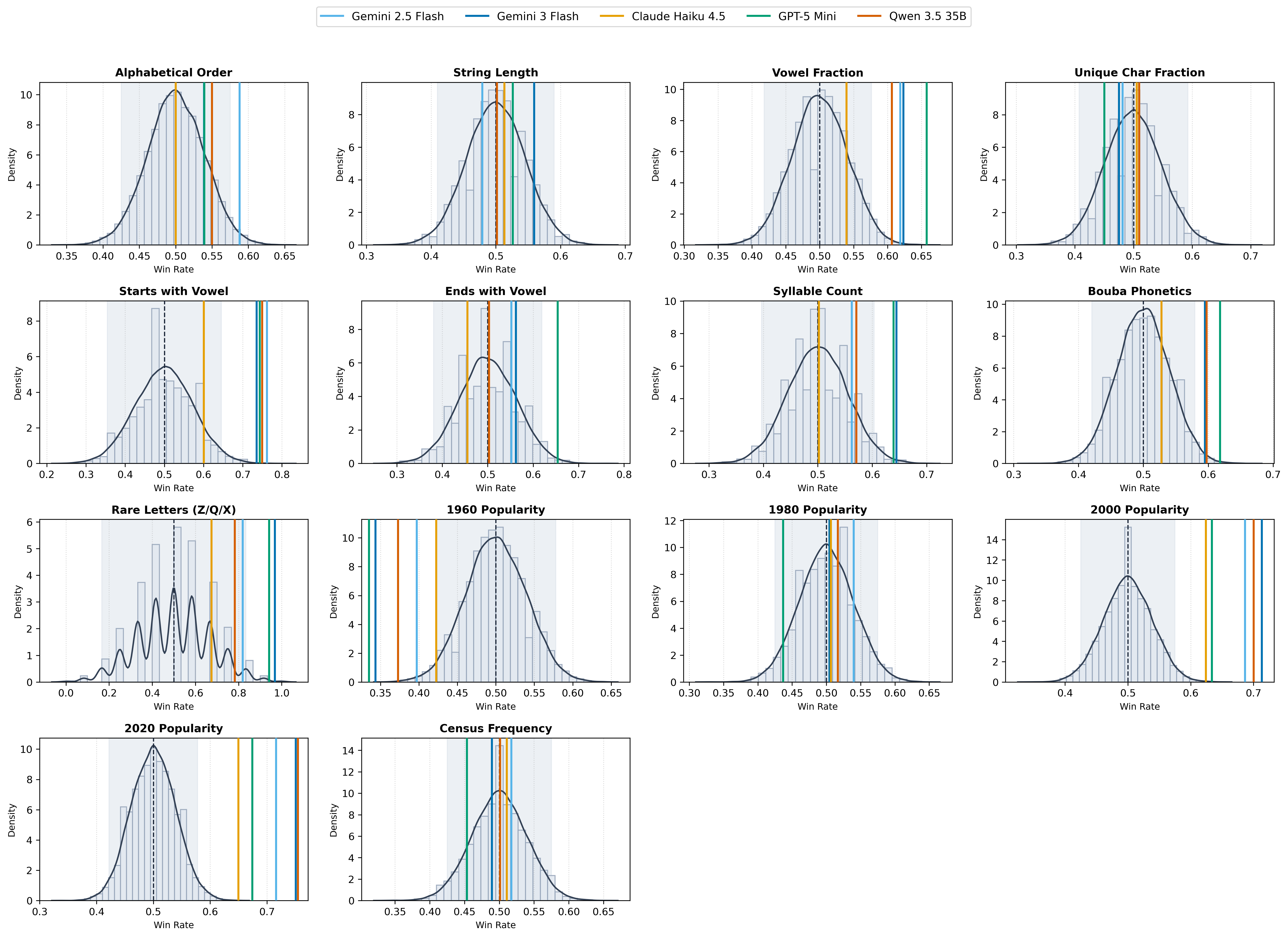}
      \caption{Names}
    \end{subfigure}
    \vspace{-4pt}
    \begin{subfigure}{0.77\linewidth}
      \centering
      \includegraphics[width=1\linewidth]{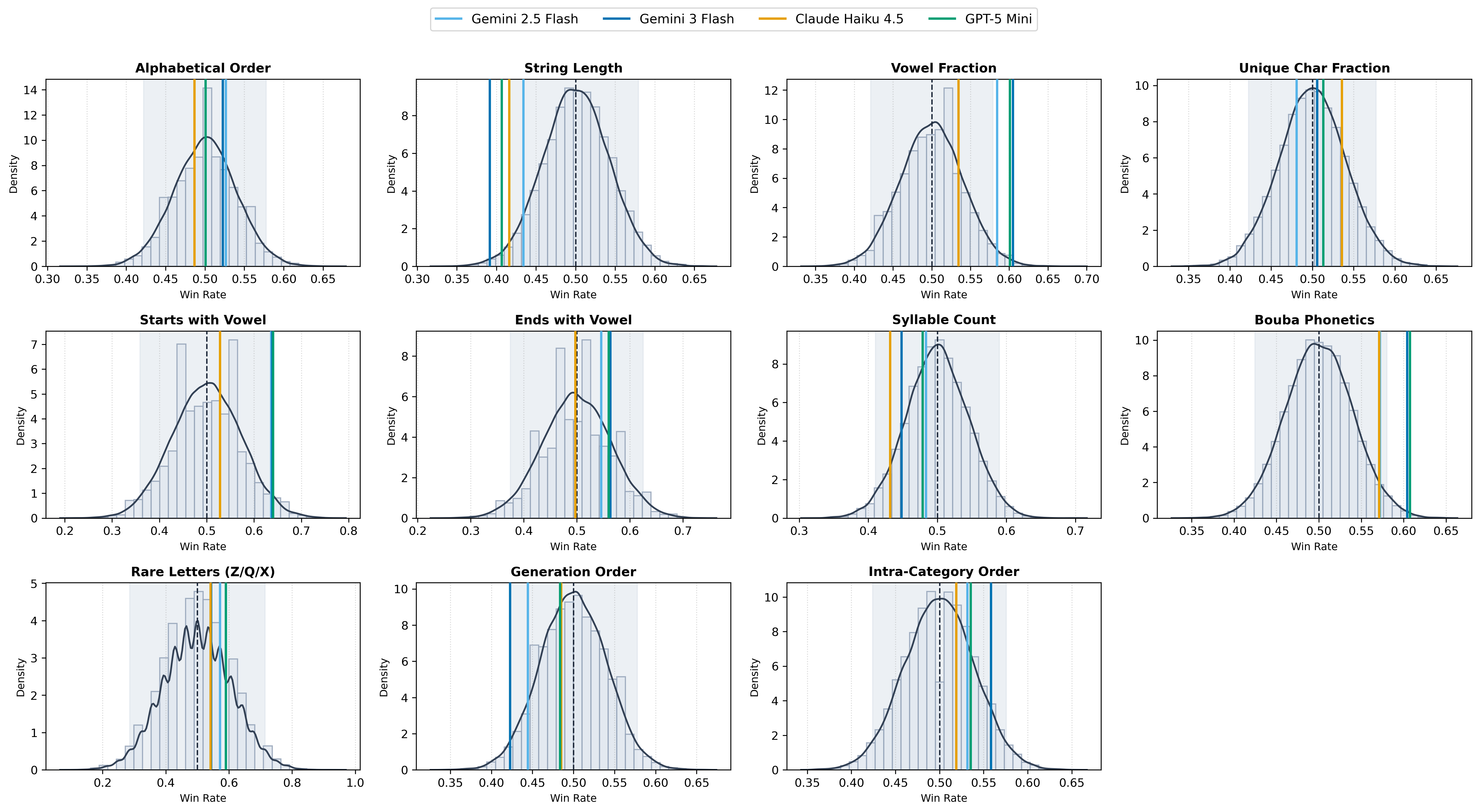}
      \caption{Cities}
    \end{subfigure}
    \caption{Monte Carlo permutation test results for the Numbers, Names, and Cities datasets. Histograms for each feature represent the win rates for that feature on 25,000 random preference permutations (see Section~\ref{app_sec:monte_carlo}), and gray shading represents 95\% confidence intervals on the Monte Carlo win rates. Each vertical line represents the win rate for that feature for a specific LLM, as in Table~\ref{tab:A.1_bias_fracs}.}
    \label{fig:app_montecarlo}
\end{figure*}

Figure \ref{fig:app_montecarlo} shows the results of this procedure. The histogram of Monte Carlo win rates is shown, and the 95\% confidence interval is shown in gray. The actual win rates observed by asking the LLMs (the values shown in Table~\ref{tab:A.1_bias_fracs}) are shown as vertical colored lines.

For more details, see Section 4 of \url{https://anon-for-reviews.github.io/neurips2026/report_statistics.html}.

\subsection{Causal Evaluation via Minimal Pairs}
\label{app_sec:minimal_pairs}
To disentangle confounding features (e.g., the correlation between "Higher" value and "Higher Sum of Digits" in the Numbers dataset), we conducted a causal evaluation using a "minimal pairs" framework. For each target feature, we generated pairs of items where one feature varied while all other identified features were held as constant as possible. To ensure spatial coverage, we used K-Means stratification to sample representative pairs across the non-target covariate space.

The learned logistic regression model fit on the original observational data (Section~\ref{sec:methods}) accurately predicts causal win rates in these experimental settings:
\begin{itemize}
    \item \textbf{Numbers Domain}: Pearson $r = 0.941$, Mean Absolute Error (MAE) = 0.051.
    \item \textbf{Names Domain}: Pearson $r = 0.884$, MAE = 0.078.
\end{itemize}

The most important preference drivers remain strong and invariant across methodologies. For instance, in the Numbers domain, the \textit{Mult 10} feature shows a causal win rate increase of +31\% to +35\% relative to a baseline of 0.5. In the Names domain, the \textit{Higher Frac of 2020 Babies} shows a causal increase of +25\% to +36\%. Full data and interactive comparisons for this causal decomposition are available at \url{https://anon-for-reviews.github.io/neurips2026/report_minimal_pairs.html}.

\subsection{City Categories: Robustness to List Order (Shuffling)}
\label{app_sec:category_shuffling}
To ensure that the endogenous categories identified in Section~\ref{sec:ideation} are not simply artifacts of the original list position, we performed a robustness check by shuffling the list of 200 cities five times. For each shuffle, we asked an LLM to re-categorize the cities, both with and without an "order hint" (the original generation rank).

The resulting categorizations showed high semantic agreement with the paper's original 19-category labeling (Gemini 3 Pro), with an Adjusted Rand Index (ARI) consistently indicating robust recovery of the thematic clusters regardless of list order (ARI range: 0.50--0.65). Full data, prompts, and interactive visualizations for these shuffled re-categorizations are available at \url{https://anon-for-reviews.github.io/neurips2026/report_categorization.html}.


\subsection{Mixed-Effects Model for Prompt Robustness}
\label{app_sec:mixed_effects}
To distinguish prompt-level variance from sampling noise, we employed a Generalized Linear Mixed-Effects Model (GLMM) with crossed random effects for both prompts and item-pairs.

Levene's test for variance comparison confirms that prompt variance significantly exceeds pure sampling noise ($p < 0.001$) for nearly all models and datasets. Variance decomposition (Table~\ref{tab:app_variance_decomp}) indicates that prompt variance typically accounts for 1--20\% of the total observed variance in win rates, while item-pair variance accounts for 3--43\%. This confirms that prompt sensitivity is a robust and significant property of model behavior, rather than an artifact of finite sampling. Full data and interactive variance decompositions for all models and datasets are available at \url{https://anon-for-reviews.github.io/neurips2026/report_statistics.html}.

\begin{table}[htb]
\centering
\caption{Variance component breakdown and Levene's test for prompt-level sensitivity. Total per-prompt win rate variance is decomposed into Prompt Variance ($\sigma^2_{\text{prompt}}$), Item-Pair Variance ($\sigma^2_{\text{pair}}$), and Sampling/Residual Noise. Levene's $W$ statistic tests whether prompt variance significantly exceeds pure sampling noise.}
\label{tab:app_variance_decomp}
\resizebox{\textwidth}{!}{
\begin{tabular}{ll ccc rr c}
\toprule
Dataset & Model & $\sigma^2_{\text{prompt}}$ (\%) & $\sigma^2_{\text{pair}}$ (\%) & Noise (\%) & $W$ Stat & $p$-val & Sig. \\
\midrule
Numbers & Gemini 2.5 Flash & 0.006 (10.2\%) & 0.003 (4.4\%) & 0.051 (85.4\%) & 120.82 & <.001 & *** \\
Numbers & Gemini 3 Flash & 0.001 (3.6\%) & 0.005 (13.4\%) & 0.032 (83.0\%) & 0.24 & .622 & n.s. \\
Numbers & Claude Haiku 4.5 & 0.010 (19.5\%) & 0.005 (9.9\%) & 0.036 (70.5\%) & 200.09 & <.001 & *** \\
Numbers & GPT-5 Mini & 0.004 (8.6\%) & 0.002 (4.1\%) & 0.043 (87.3\%) & 7.36 & .007 & * \\
Numbers & Qwen 3.5 35B & 0.003 (4.5\%) & 0.003 (4.6\%) & 0.051 (90.9\%) & 18.85 & <.001 & *** \\
\midrule
Names & Gemini 2.5 Flash & 0.003 (4.8\%) & 0.002 (3.2\%) & 0.059 (92.0\%) & 27.99 & <.001 & *** \\
Names & Gemini 3 Flash & 0.002 (5.0\%) & 0.006 (16.4\%) & 0.029 (78.6\%) & 3.01 & .083 & n.s. \\
Names & Claude Haiku 4.5 & 0.004 (6.7\%) & 0.026 (42.2\%) & 0.031 (51.1\%) & 19.02 & <.001 & *** \\
Names & GPT-5 Mini & 0.005 (9.5\%) & 0.005 (9.5\%) & 0.040 (81.0\%) & 5.94 & .015 & * \\
Names & Qwen 3.5 35B & 0.004 (7.2\%) & 0.006 (9.3\%) & 0.050 (83.5\%) & 55.27 & <.001 & *** \\
\midrule
Cities & Gemini 2.5 Flash & 0.002 (3.0\%) & 0.004 (5.6\%) & 0.059 (91.4\%) & 11.25 & <.001 & *** \\
Cities & Gemini 3 Flash & 0.001 (1.3\%) & 0.011 (27.7\%) & 0.029 (71.0\%) & 7.45 & .007 & * \\
Cities & Claude Haiku 4.5 & 0.011 (16.2\%) & 0.012 (17.0\%) & 0.045 (66.8\%) & 116.17 & <.001 & *** \\
Cities & GPT-5 Mini & 0.004 (7.1\%) & 0.006 (11.0\%) & 0.047 (81.9\%) & 16.84 & <.001 & *** \\
\bottomrule
\end{tabular}
}
\end{table}

\subsection{Model Consistency: Feedback Arc Sets} \label{sec:consistency}
We ask now whether the LLM provides \emph{consistency} in its comparisons. In a perfectly consistent setting, the items could be ordered from best to worst, and each comparison question to the LLM would return a result favoring the better of the two options. We study the degree of consistency as follows. Given $N$ items, there are $2\binom{N}{2}$ possible ordered pairs. We may ask the model to produce a comparison of each such ordered pair, and add an edge to a multigraph for each outcome directed from the winner to the loser. For a perfectly consistent judge, the vertices of the graph can be numbered so that every edge goes from a lower number to a higher number. For inconsistent judges, though, we can number the vertices so that as many edges as possible go from a lower number to a higher number; the remaining edges that go from nominally worse to better items are called the \emph{Minimum Feedback Arc Set (FAS)} of the graph, and the number of such edges is a measure of the inconsistency of the judge.

We compute the sizes of these sets in two settings. First, we consider a ``wisdom of crowds'' approach in which, for every ordered pair, each prompt votes for a winner and the majority prevails. Second, we consider a per-prompt setting in which each individual prompt considers each pair in each order, compares them, and outputs the winner. The approximate minimum FAS is calculated using the techniques of \citet{eades1993fast}. For comparison, we also calculate feedback arc sets for 50 random graphs (same edge set, but edge directions determined randomly).

\begin{figure}[htb]
    \centering
    \includegraphics[width=0.6\linewidth]{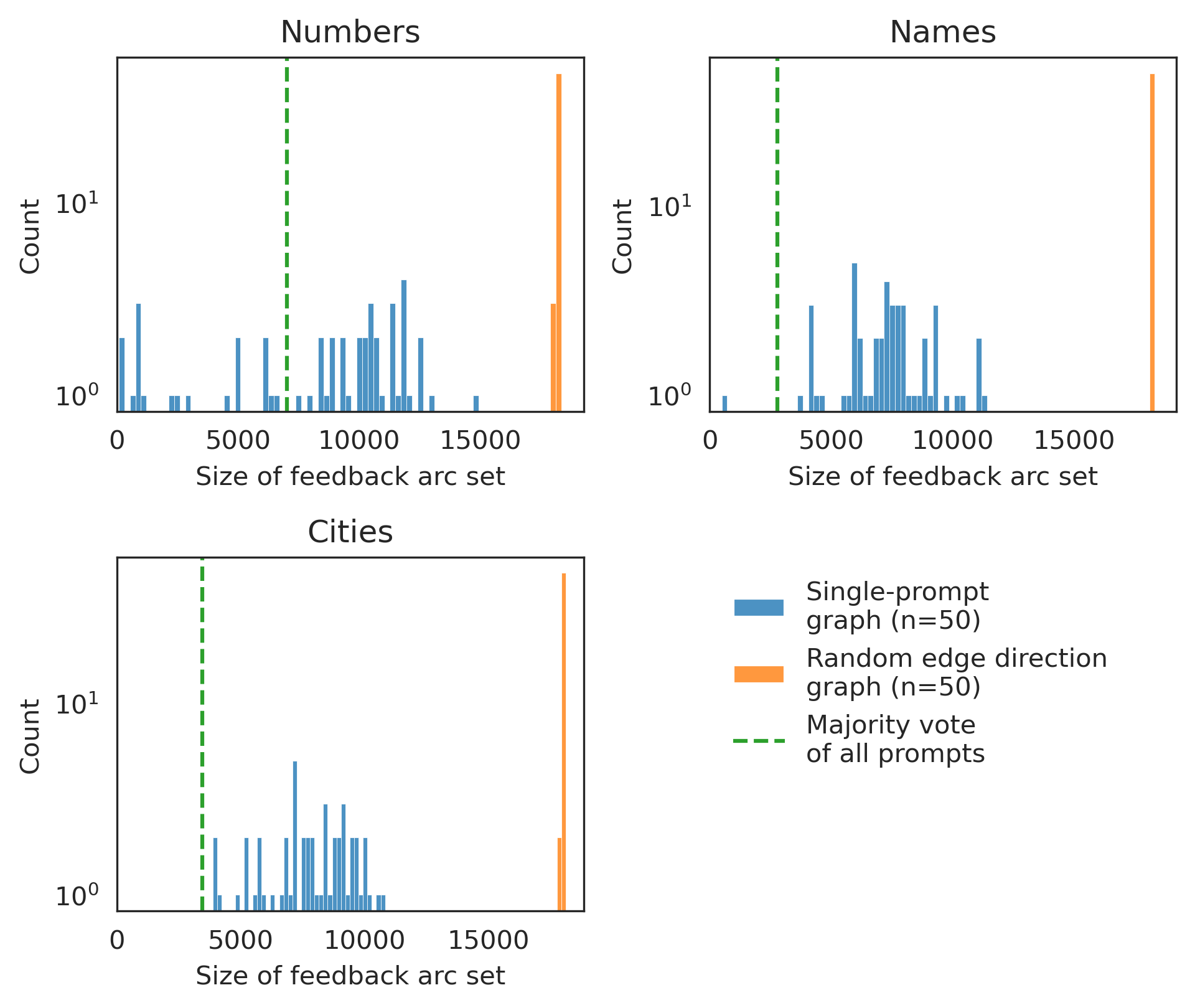}
    \caption{Model consistency: Sizes of minimum feedback arc sets (FAS) for directed preference graphs. The dotted line (green) shows the FAS size for the directed graph formed by majority vote over 50 prompts. Also shown are FAS sizes for each individual prompt's directed graph (blue) and for 50 graphs with random edge directions (orange).}
    \label{fig:C_graph_fas}
\end{figure}

Figure~\ref{fig:C_graph_fas} shows the sizes of the FAS for both settings. For the wisdom of crowds approach combining all prompts, the FAS for numbers contains 17\% of the edges, while for cities and names the FAS contains 9\% and 7\% of the edges respectively. Thus, the cross-prompt consensus formed in the numbers domain is only mildly consistent, while in the names domain it is strongly consistent. We note that this consistency analysis was performed on an older dataset with a different combinatorial design (every possible pair evaluated once in each order, rather than the random cycle pairing used in Section~\ref{sec:methods}); for names, this older dataset also used an LLM-generated name list rather than the census-based list used in Section~\ref{sec:bias}.

The per-prompt FAS sizes, on the other hand, span a wider range, especially for numbers. Every prompt provides judgements that are more consistent than random guessing. However, presentation order effects make the individual judgements inherently less consistent. For names, the per-prompt FASes contain anywhere from 10--28\% of the edges (except for a single outlier prompt that preferred alphabetic order almost all the time). The two most inconsistent prompts for names ($\approx 28\%$ edges) are: ``Of the two names, which one would you personally endorse?'' (Prompt 31) and ``Please identify your preferred option among these two names.'' (Prompt 18).

For numbers, the situation is even more varied. Some prompts are almost perfectly consistent (Prompt 23, ``Regarding these two numbers, which one would you put first?'', has a FAS of only 0.2\% of edges), while others are highly inconsistent (Prompt 46, ``Which of the two numbers do you personally fancy?'', has a FAS containing 1/3 of the edges). In conclusion, small prompt variations result in differences not just in outcomes, but in LLM reproducibility.

\clearpage

\section{Comparable objects}
\label{app_sec:comp_object_tables}



\subsection{Cities and categories}
See Table~\ref{tab:J_city_cats}. Categorization was produced by an LLM (Prompt~\ref{prompt:city_categorization}) that had no context on the process for generating the city names.

\begin{table*}[p]
    \centering
    \fontsize{8pt}{9pt}\selectfont
    \setlength{\tabcolsep}{4pt}
    \renewcommand{\arraystretch}{0.92}
\begin{tabularx}{\textwidth}{>{\raggedright\arraybackslash}p{2.6cm} X}
\toprule
Category (range) & Cities \\
\midrule
Landscape and Direction \newline (1--28) & Silverglade, Veridian Bay, Whispering Pines, Azure Peak, Crimson Ridge, Dragon's Tooth, Shadowfen, Ironwood, Starfall, Sunstone, Mooncrest, Emberfall, Stormwind, Frostholm, Greenfield, Oakheart, Serpent's Coil, Griffin's Roost, Titan's End, Havenwood, Northridge, Southgate, Westwind, Eastbrook, Riverbend, Lakewatch, Seacliff, Stonehaven \\
\midrule
Precious Metals \newline (29--34) & Goldcrest, Silvercreek, Bronzeleaf, Copperhead, Ironhand, Steelheart \\
\midrule
Gemstones \newline (35--49) & Diamondfall, Rubyvale, Emerald City, Sapphire Isle, Amethyst Cove, Topaz Point, Obsidian Shore, Onyx Ridge, Pearl Bay, Coral Gables, Jasper Dale, Garnet Grove, Opal Springs, Turquoise River, Jade Mountain \\
\midrule
Elevation Terms \newline (50--57) & Zenith, Nadir, Apex, Pinnacle, Summit, Crestfall, Valedale, Hilltop \\
\midrule
Hydrology \newline (58--62) & Clearwater, Blackwater, Whitewater, Redwater, Bluewater \\
\midrule
Field Colors \newline (63--65) & Greenfield, Brownfield, Greyfield \\
\midrule
Seasons \newline (66--69) & Springdale, Summerwind, Autumnwood, Winterfell \\
\midrule
Nobility \newline (70--80) & Highgarden, Kingswood, Queensgate, Prince's Landing, Princess Anne's Harbor, Duke's Pass, Duchess's Crossing, Baron's End, Baroness's View, Knight's Watch, Squire's Rest \\
\midrule
Military Units \newline (81--85) & Archer's Point, Bowman's Ford, Swordsman's Creek, Spearman's Ridge, Axeman's Hollow \\
\midrule
The Forge \newline (86--89) & Hammerfall, Anvil Rock, Forgefire, Smeltertown \\
\midrule
Medieval Trades \newline (90--107) & Miner's Deep, Farmer's Field, Shepherd's Crook, Fisherman's Wharf, Hunter's Lodge, Trapper's Cabin, Lumberjack's Mill, Blacksmith's Forge, Carpenter's Bench, Mason's Quarry, Weaver's Loom, Potter's Wheel, Chandler's Wick, Cooper's Barrel, Fletcher's Arrow, Wainwright's Wheel, Cartwright's Path, Shipwright's Harbor \\
\midrule
Arts and Sciences \newline (108--113) & Glassblower's Furnace, Alchemist's Tower, Scribe's Library, Bard's Tale, Minstrel's Song, Jester's Court \\
\midrule
Chess Strategy \newline (114--119) & King's Ransom, Queen's Gambit, Bishop's Crossing, Rook's End, Pawn's Opening, Checkmate \\
\midrule
Mythical Beasts \newline (120--143) & Dragon's Breath, Griffin's Claw, Unicorn's Horn, Phoenix's Ashes, Wyvern's Sting, Kraken's Deep, Leviathan's Trench, Behemoth's Gate, Manticore's Lair, Harpy's Nest, Centaur's Grove, Satyr's Glade, Nymph's Spring, Dryad's Oak, Pixie's Hollow, Gnome's Grotto, Dwarf's Mine, Elf's Forest, Orc's Stronghold, Goblin's Cave, Troll's Bridge, Ogre's Swamp, Giant's Causeway, Titan's Rise \\
\midrule
Mythology \newline (144--151) & Olympus, Asgard, Valhalla, Elysium, Avalon, Shangri-La, El Dorado, Atlantis \\
\midrule
Ancient Cities \newline (152--161) & Babylon, Camelot, Carthage, Constantinople, Alexandria, Rome, Athens, Sparta, Thebes, Troy \\
\midrule
Fantasy Literature \newline (162--170) & Valyria, Westeros, Essos, Middle-earth, Narnia, Oz, Neverland, Wonderland, Fantasia \\
\midrule
Fantasy Regions \newline (171--182) & Aethelgard, Brimstone, Cedarwood, Duskwood, Evermore, Farshore, Grimwade, Hollowdale, Islesmere, Silverwood, Port Anson, Brighthaven \\
\midrule
The Sims 4 Worlds~\citep{sims4wiki} \newline (183--200) & Newcrest, Willow Creek, Oasis Springs, Magnolia Promenade, Granite Falls, Windenburg, San Myshuno, Forgotten Hollow, Brindleton Bay, Del Sol Valley, Strangerville, Sulani, Glimmerbrook, Britechester, Evergreen Harbor, Mt. Komorebi, Henford-on-Bagley, Copperdale \\
\bottomrule
\end{tabularx}

\caption{City names and categories. Ranges give each category's positions in the original generated list; names are presented in the order they appeared in the list.}
\label{tab:J_city_cats}
\end{table*}

\clearpage

\section{Prompt questions for comparisons: set of 50}
\label{app_sec:prompt_tables}
Table~\ref{tab:M_prompts}, below, lists all comparison question prompts and the relevant categories assigned by Gemini. Tables ~\ref{tab:prompt_cat_grammatical_explanation} and~\ref{tab:prompt_cat_semantic_domain} show the definition Gemini gave for each category. Categories were invented by Gemini when asked to categorize the list of prompts by grammatical structures.


\footnotesize
\begin{xltabular}{\textwidth}{r X l l l}
\caption{Comparison question prompts used for all experiments. See Tables~\ref{tab:prompt_cat_grammatical_explanation} and~\ref{tab:prompt_cat_semantic_domain} for definitions of Ling(uistic) cat(egory) and Sem(antic) dom(ain).}
\label{tab:M_prompts}
\\
\toprule
Prompt & Prompt & Ling. & Sem. & Tone \\
index & & cat. & dom. & \\
\midrule
\endfirsthead
\multicolumn{5}{l}{\textit{Table~\ref{tab:M_prompts} (continued)}} \\
\toprule
Prompt & Prompt & Ling. & Sem. & Tone \\
index & & cat. & dom. & \\
\midrule
\endhead
\bottomrule
\endfoot
0 & What is your preference between these two \{numbers,names,cities\}? & SI & D & Direct \\
1 & Between these two \{numbers,names,cities\}, which do you favor? & F & J & Direct \\
2 & Which of the two \{numbers,names,cities\} do you find more appealing? & SI & A & Direct \\
3 & If you had to choose between these two \{numbers,names,cities\}, which would it be? & C & D & Soft \\
4 & Do you have a stronger inclination towards one of these two \{numbers,names,cities\}? & SI & P & Soft \\
5 & Which \{number,name,city\} out of these two options would you select? & P & D & Direct \\
6 & Out of the two provided \{numbers,names,cities\}, which one do you like better? & F & P & Direct \\
7 & Please indicate which of these two \{numbers,names,cities\} you lean towards. & I & P & Formal \\
8 & Which of these two \{numbers,names,cities\}, in your opinion, is the better choice? & P & J & Soft \\
9 & Considering these two \{numbers,names,cities\}, which one resonates more with you? & F & A & Soft \\
10 & Given these two \{numbers,names,cities\}, which one would you rather select? & F & D & Direct \\
11 & When considering these two \{numbers,names,cities\}, which do you see as more desirable? & C & A & Formal \\
12 & Out of the current pair of \{numbers,names,cities\}, which is your top choice? & F & D & Direct \\
13 & Tell me, which of these two \{numbers,names,cities\} do you prefer over the other? & I & D & Direct \\
14 & Which \{number,name,city\}, from the duo presented, holds more appeal for you? & P & A & Formal \\
15 & I'd like to know which of these two \{numbers,names,cities\} you find preferable. & D & J & Soft \\
16 & If you were to rate these two \{numbers,names,cities\}, which would score higher in your estimation? & C & J & Formal \\
17 & Between these two \{numbers,names,cities\}, which do you hold in higher regard? & F & J & Formal \\
18 & Please identify your preferred option among these two \{numbers,names,cities\}. & I & D & Formal \\
19 & Which \{number,name,city\}, when weighed against the other, would you ultimately favor? & P & J & Formal \\
20 & Which \{number,name,city\}, from the pair, gets your vote? & P & D & Casual \\
21 & Among the two \{numbers,names,cities\} presented, which one stands out to you? & F & A & Direct \\
22 & Given these two options, which \{number,name,city\} would you opt for? & F & D & Direct \\
23 & Regarding these two \{numbers,names,cities\}, which one would you put first? & F & D & Direct \\
24 & Between the two \{numbers,names,cities\} provided, which one earns your approval? & F & J & Formal \\
25 & Which of these two \{numbers,names,cities\} would you single out? & SI & D & Direct \\
26 & Which of these two \{numbers,names,cities\} would you champion? & SI & D & Direct \\
27 & Which of these two \{numbers,names,cities\} would you rather go with? & SI & D & Direct \\
28 & What is your D choice between these two \{numbers,names,cities\}? & SI & D & Formal \\
29 & Which \{number,name,city\}, given this duo, would you ultimately settle on? & P & D & Direct \\
30 & Which of these two \{numbers,names,cities\} do you find more agreeable? & SI & P & Soft \\
31 & Of the two \{numbers,names,cities\}, which one would you personally endorse? & F & J & Formal \\
32 & Between the two \{numbers,names,cities\}, which makes a more positive impression on you? & F & P & Soft \\
33 & Regarding these two \{numbers,names,cities\}, which one holds a higher subjective value for you? & F & J & Formal \\
34 & Which \{number,name,city\}, from this pair, do you perceive as more favorable? & P & J & Formal \\
35 & Out of these two \{numbers,names,cities\}, which one speaks more to your sensibility? & F & A & Soft \\
36 & Which of the two \{numbers,names,cities\} do you feel a greater affinity for? & SI & P & Soft \\
37 & Which of the two \{numbers,names,cities\} do you deem more fitting? & SI & J & Formal \\
38 & For these two \{numbers,names,cities\}, which one commands your strongest positive reaction? & F & P & Formal \\
39 & Which \{number,name,city\}, from the presented duo, do you find yourself drawn to more? & P & P & Soft \\
40 & Which of these two \{numbers,names,cities\} do you find more commendable? & SI & J & Formal \\
41 & Between these two \{numbers,names,cities\}, which elicits a more positive internal response from you? & F & P & Formal \\
42 & Which \{number,name,city\}, when presented side-by-side, do you find more compelling? & P & A & Direct \\
43 & Regarding this duo, which \{number,name,city\} truly captures your imagination? & F & A & Soft \\
44 & Which of these two \{numbers,names,cities\} do you consider to have a greater intrinsic value? & SI & J & Formal \\
45 & Which \{number,name,city\} from this selection of two do you anticipate liking more? & SI & P & Soft \\
46 & Which of the two \{numbers,names,cities\} do you personally fancy? & SI & P & Casual \\
47 & From the pair provided, which \{number,name,city\} generates a more positive aesthetic response in you? & F & A & Formal \\
48 & Which of these two \{numbers,names,cities\} truly captivates your interest? & SI & P & Soft \\
49 & Considering these two \{numbers,names,cities\}, which one would you most readily adopt? & F & D & Direct \\
\end{xltabular}

\begin{table*}
\footnotesize
\centering
\begin{tabularx}{\textwidth}{l l X X}
\toprule
\textbf{Category} & \textbf{Abbrv.} & \textbf{Movement/Grammar Logic} & \textbf{Resulting Tone} \\
\midrule
Conditional & C & Uses "If/When" clauses to set a hypothesis. & Softened; polite; indirect. \\
Declarative & D & Formulated as a statement rather than a question. & Indirect; gentle; observational. \\
Fronting & F & Moves a phrase (often prepositional) to the start. & Analytical; context-first. \\
Imperative & I & Uses a command verb (e.g., "Please indicate"). & Instructional; administrative. \\
Parenthetical & P & Inserts a qualifying phrase inside the sentence. & Nuanced; deliberate; subjective. \\
Standard Inversion & SI & Flips Subject and Auxiliary verb for a question. & Direct; neutral; efficient. \\
\bottomrule
\end{tabularx}
\caption{\enquote{Linguistic categories} as defined by Gemini. Generated in a conversation about categorizing the prompts in Table \ref{tab:J_city_cats}  by grammatical structures.}
\label{tab:prompt_cat_grammatical_explanation}
\end{table*}

\begin{table*}[h]
\footnotesize
\centering
\begin{tabularx}{\textwidth}{l l X X}
\toprule
\textbf{Category} & \textbf{Abbrv.} & \textbf{Key Words Used} & \textbf{Resulting Tone} \\
\midrule
Aesthetic & A & Appealing, resonates, aesthetic response, imagination & Soft, artistic, and subjective. \\
Decisive & D & Opt for, single out, settle on, champion & Action-oriented and final. \\
Judgmental & J & Deem, fitting, commendable, endorse, value & Serious, objective, and weighty. \\
Psychological & P & Affinity, inclination, sensibility, internal response & Analytical and introspective. \\
\bottomrule
\end{tabularx}
\caption{\enquote{Semantic domain}, as defined by Gemini: \enquote{four distinct \enquote{vocal styles} based on the verbs used}. Generated in a conversation about categorizing the prompts in Table \ref{tab:J_city_cats}  by grammatical structures.}
\label{tab:prompt_cat_semantic_domain}
\end{table*}

\clearpage

\section{Other prompts used} \label{app_sec:prompts_used}
Unless otherwise specified, all prompts were used with Gemini 2.5 Flash, using the standard Gemini API.

For the comparison question prompts (the ``{{question prompt}}'' in the template in Appendix~\ref{app_sec:comparison_prompt_template}), see Appendix~\ref{app_sec:prompt_tables}.

\begin{prompt}
    \begin{promptbox}
    \begin{lstlisting}
You are an expert prompt engineer. You need to generate 10
    new prompts that represent different ways to ask a model for the same
    thing. The base prompt you are supposed to expand is: Which of these two names do you prefer?.
    (*\choice{\{,In addition to the base prompt, another model has} \choice{already produced an expanded set of prompts } \choice{listed here:}

       
       \choice{<begin generated prompts>} \choice{
      \var{\{\{already\_generated\_prompts\}\}}} \choice{
      <end generated prompts>\}}*)
      Please output 10 new prompts, one per line, such that
    each new prompt has the same meaning as the base prompt. Do not number
    the prompts; simply output each on its own line. Each new
    prompt should mean the same thing as the base prompt. However, each new
    prompt should be written differently from the base prompt, the
    expanded prompts, and any new generated prompts you have
    already produced.
    \end{lstlisting}
 \end{promptbox}
 \caption{Expansion for set of 50 prompts.}
\label{prompt:prompt_expander}
\end{prompt}

\begin{prompt}
    \begin{promptbox}
\begin{lstlisting}
(*\user*)Here is a list of city names. Please fill in 'Category Name' with an appropriate category. Keep the set of categories to 8-12 categories, and try to keep the sizes of the categories balanced. However, prioritize in this order: clear, meaningful, well-separated categories; size balance; using any meaning found in the order of the list; anything else. all_city_cat_lists['g3P04'] =  [
    ['Silverglade', 'Category Name'],
    ['Veridian Bay', 'Category Name'],
    (*\var{\{\{rest of city names\}\}}*)

(*\model*)Here is the Python list with the filled categories. I have identified 9 distinct categories [...]

(*\user*)Would the list be different if the requirements for category count and list ordering were relaxed?
    
 \end{lstlisting}
 \end{promptbox}
 \caption{Labeling of city categories in Table ~\ref{tab:J_city_cats}. Prompt was given to Gemini 3 Pro. (Note that order of the list \textbf{is} mentioned in the prompt. However, this is because when not given instructions on priorities, the model tended to prioritize order of the list over everything else.)}
\label{prompt:city_categorization}
\end{prompt}

\subsection{Subjective Preference Comparison Prompt Template} \label{app_sec:comparison_prompt_template}
To ask the model to choose between two candidates \texttt{ent1} and \texttt{ent2} using a given prompt phrasing \texttt{prompt}, we construct a standardized preference comparison template:

\begin{prompt}
\begin{promptbox}
\begin{lstlisting}
(*\var{\{\{question prompt\}\}}*)
The two numbers are: [(*\var{\{\{ent1\}\}}*)] | [(*\var{\{\{ent2\}\}}*)].
Please respond with only the chosen number.
\end{lstlisting}
\end{promptbox}
\caption{Standardized preference comparison template used for numbers.}
\label{prompt:comparison_template_numbers}
\end{prompt}

\begin{prompt}
\begin{promptbox}
\begin{lstlisting}
(*\var{\{\{question prompt\}\}}*)
The two names are: [(*\var{\{\{ent1\}\}}*)] | [(*\var{\{\{ent2\}\}}*)].
Please respond with only the chosen name.
\end{lstlisting}
\end{promptbox}
\caption{Standardized preference comparison template used for human names.}
\label{prompt:comparison_template_names}
\end{prompt}

\begin{prompt}
\begin{promptbox}
\begin{lstlisting}
(*\var{\{\{question prompt\}\}}*)
The two cities are: [(*\var{\{\{ent1\}\}}*)] | [(*\var{\{\{ent2\}\}}*)].
Please respond with only the chosen city.
\end{lstlisting}
\end{promptbox}
\caption{Standardized preference comparison template used for fictional city names.}
\label{prompt:comparison_template_cities}
\end{prompt}

\begin{prompt}
    \begin{promptbox}
\begin{lstlisting}
From the AI model output below, identify and list the key factors that significantly influenced its decision-making process. Present these factors as two Python-style lists of strings (e.g., ["Prime", "Larger", "Odd", "Even", "First presented", "Repeated digits"], ["Larger", "Prime"]). The first list should contain all factors mentioned in the model output, in the order they appear, represented using the fewest words possible. The second list should contain only the factors demonstrably used in the final decision, ordered by their relative importance, with the most influential factors listed first. This second list should only include factors that are directly relevant to the option that was ultimately chosen - e.g. "Larger" or "Round" but not "Deterministic choices" or "Predictable". The factors in the second list should also describe the option that was ultimately chosen - e.g. if it was deciding between a larger and smaller number, and ultimately chose the smaller one, the second list should contain "Smaller" but not "Larger". Factors should be represented as concise, one-to-three word phrases, using the fewest words possible while keeping meaning clear - e.g "Larger" instead of "Larger number". Ensure consistent terminology across factors. Only factors relevant to the inputs should be included - exclude statements of neutrality (e.g., "no preference") or limitations (e.g., "lack guidance"). If no relevant factors are identified in the model output, state ["No factors identified"] for both lists. Base the lists solely on the provided model output; do not introduce external information or assumptions. The output should contain *only* the two lists, with no additional text or formatting. (*\var{\{\{model\_output\}\}}*)
\end{lstlisting}
 \end{promptbox}
 \caption{Getting list of most important decision factors for each thinking trace.}
 \label{prompt:thinking_analysis}
\end{prompt}

\clearpage


\section{Summary Table of Preference Features} \label{app:bias-list}

{\footnotesize
\begin{tabularx}{\textwidth}{l l X}
\toprule
Feature & Datasets & Notes \\
\midrule
Asc Digit Order & Numbers & Digits are in ascending order (smallest number first). \\
Desc Digit Order & Numbers & Digits are in descending order (largest number first). \\
Ending Vowel & Cities and names & Ends with a vowel. \\
First Alphabetically & Cities and names &  \\
First Generated & Cities & Earlier in the list of entities originally generated by an LLM. \\
First Presented & All domains & First option shown to the model. \\
Generated Earlier Within Own Category & Cities & Ranked higher in percentile rank (each within their own category). \\
Has Repeat Digits & Numbers & Digits are not unique (e.g. 121 or 844). \\
Higher & Numbers & Higher number. \\
Higher Bouba Frac & Cities and names & Higher fraction of `bouba vs. ``kiki letters (see text for citation). \\
Higher Census Freq & Names & Higher frequency on 2020 US Census (see text for citation) \\
Higher Frac of 1960 Babies & Names & Name was more common among babies born in 1960. \\
Higher Frac of 1980 Babies & Names & See prev \\
Higher Frac of 2000 Babies & Names & See prev \\
Higher Frac of 2020 Babies & Names & See prev \\
Higher Order Mag & Numbers & Higher order of magnitude. \\
Higher Rare Letter Frac & Cities and names & Higher fraction of ``rare letters (LLM-defined as Q, X, or V). \\
Higher Sum Digits & Numbers & Sum of number's digits is higher. \\
Higher Syllable Heur & Cities and names & Higher syllable count (using a heuristic based on vowels). \\
Higher Unique Char Frac & Cities and names & Higher ratio of unique letter count to total letter count. \\
Higher Vowel Frac & Cities and names & Higher ratio of vowel count to total letter count. \\
Longer Length & Cities and names & Longer string. \\
Mult 10 & Numbers & Multiple of 10. \\
Mult 2 & Numbers & See prev \\
Mult 5 & Numbers & See prev \\
Palindrome & Numbers &  \\
Prime & Numbers &  \\
Starting Vowel & Cities and names & Starts with a vowel. \\
\bottomrule
\end{tabularx}

}

\newpage
\section{Model Configurations and Citations} \label{app_sec:model_configs}
Below, we provide a list of all model families used in our experiments along with their exact API, HuggingFace, or LiteLLM identifiers and their respective citations.

\begin{table}[htbp]
\centering
\footnotesize
\begin{tabular}{llll}
\toprule
\textbf{Creator} & \textbf{Model Family} & \textbf{API / Config String} & \textbf{Citation} \\
\midrule
Google & Gemini 3 Flash & \texttt{"gemini-3-flash-preview"} & \cite{modelset_gdm} \\
Google & Gemini 2.5 Flash & \texttt{"gemini-2.5-flash-preview"} & \cite{modelset_gdmorig} \\
Qwen & Qwen 3.5 35B & \texttt{Qwen/Qwen3.5-35B-A3B} & \cite{modelset_qwen} \\
OpenAI & GPT-5 Mini & \texttt{gpt-5-mini-2025-08-07} & \cite{modelset_openai} \\
Anthropic & Claude Haiku 4.5 & \texttt{claude-haiku-4-5-20251001} & \cite{modelset_anthropic} \\
\bottomrule
\end{tabular}
\caption{Model configurations and their respective citations.}
\label{tab:model_configs}
\end{table}

\section{Factors named in thinking traces}
\label{app_sec:thinking_names}

The thinking traces for each comparison were fed to a second LLM, which was asked to generate a list of factors mentioned in the thinking, ordered by highest importance to the final decision (Prompt~\ref{prompt:thinking_analysis}). We note that due to temporary API limitations at the time of collection, this analysis for both datasets was performed on a separate evaluation set sampled from an older (LLM-generated) version of the names dataset, rather than the name list and pairing design used for the win fractions in Section~\ref{sec:bias}.

\begin{figure*}[htb]
    \centering
    \includegraphics[width=0.7\linewidth]{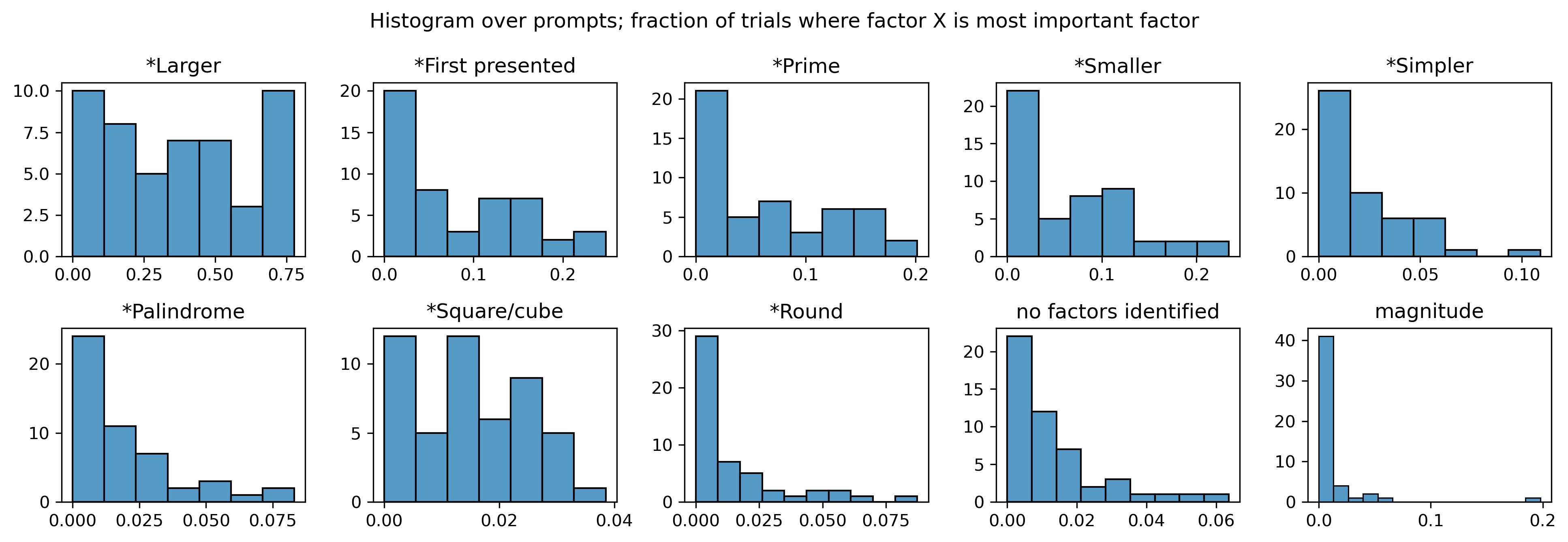}
    \caption{For each prompt, the fraction of trials where the given factor was determined from the model's thinking trace to be the most relevant decision factor for the Numbers dataset.}
    \label{fig:F_thinking_hist_numbers}
\end{figure*}

For the Numbers dataset, the most important factors identified in these traces (after manual deduplication of factor names) are shown in Figure~\ref{fig:F_thinking_hist_numbers}.

\begin{figure*}[htbp]
    \centering
    \includegraphics[width=0.7\linewidth]{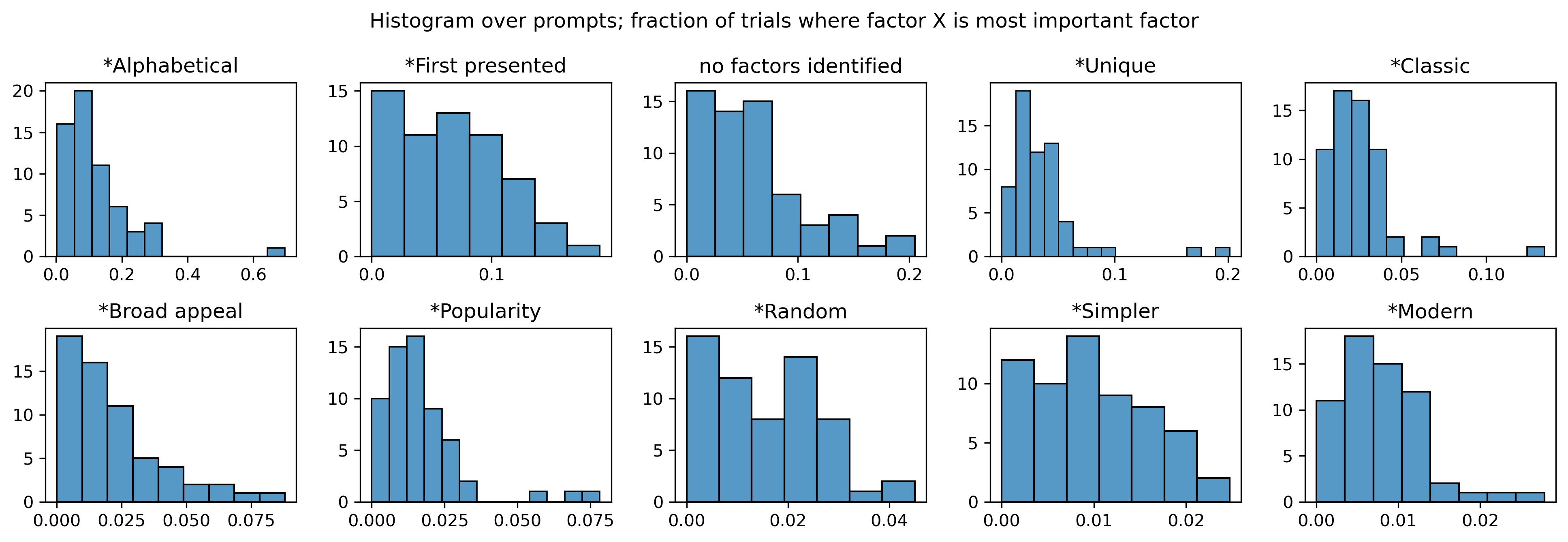}
    \caption{Histogram over prompts: for each prompt, the fraction of trials where the given factor was determined from the model's thinking trace to be the most relevant decision factor for the Names dataset.}
    \label{fig:F_thinking_hist_names}
\end{figure*}

For the Names dataset, this analysis was performed on a set of 200 unique pairs due to collection-time constraints. As shown in Figure~\ref{fig:F_thinking_hist_names}, the model's reasoning for names is significantly more complex than for numbers, frequently citing phonetics, cultural associations, and perceived ``modernity'' as decisive factors.

\section{Broader Impacts} \label{app_sec:broader_impacts}


We study how LLMs engage in subjective decision-making tasks, an activity that is used for a set of broader tasks ranging from creative brainstorming to marketing and branding. Given the widespread use of this type of task, it is important to understand how the underlying preferences in the models affect the outcomes. This can both inform users of how to interpret the results they from LLMs for these tasks, and also suggest lines of research to mitigate the biases.

We have also found that the set of preferences is not monolithic, but can in fact be affected by subtle variations in the prompts that are used to specify the subjective comparisons. We have focused on comparisons between short pieces of content, but our methodology for formulating and evaluating the comparisons provides a template that can be applied essentially directly to a wide range of other subjective comparisons. In particular, it indicates the benefits of using multiple heterogeneous prompts for these types of comparison tasks, to offset the effect of preferences inherent to any one particular prompt or style of prompt. Our empirical results provide guidelines for some of the general effects that different prompts induce on the underlying comparison activity, in general terms.

\newpage
\section{Extended Results for Other Model Families, Datasets, and Model Hyperparameters} \label{app_sec:other_models_figs}

\clearpage
\subsection{Tables of Results for Other Datasets and Model Settings} \label{app:detailed-tables}

Table~\ref{tab:A.2_bias_fracs_temperatures} lists the win fractions across multiple temperatures ($T=0.1, 0.7, 1.0$) for several different models (the selected GPT model does not permit temperatures other than 1.0).

Table~\ref{tab:A.3_bias_fracs_cities} lists win fractions for the Cities dataset, as Table~\ref{tab:A.1_bias_fracs} does for Numbers and Names.

\begin{table}[h]
\centering
\caption{Table of biases across different temperatures for each model.}
\label{tab:A.2_bias_fracs_temperatures}
\resizebox{\textwidth}{!}{
\begin{tabular}{llllllll||lll}
\toprule
 &  & \shortstack{Gemini \\ 2.5 Flash \\ (T=1.0)} & \shortstack{Gemini \\ 2.5 Flash \\ (T=0.7)} & \shortstack{Gemini \\ 2.5 Flash \\ (T=0.1)} & \shortstack{Gemini \\ 3 Flash \\ (T=1.0)} & \shortstack{Gemini \\ 3 Flash \\ (T=0.7)} & \shortstack{Gemini \\ 3 Flash \\ (T=0.1)} & \shortstack{Claude \\ Haiku 4.5 \\ (T=1.0)} & \shortstack{Claude \\ Haiku 4.5 \\ (T=0.7)} & \shortstack{Claude \\ Haiku 4.5 \\ (T=0.1)} \\
Comparable & Metric &  &  &  &  &  &  &  &  &  \\
\midrule
\multirow[c]{12}{*}{Numbers} & Higher & {\cellcolor{white}} \color{black} {\cellcolor[HTML]{3B4CC0}} \color[HTML]{F1F1F1} 0.70\textsuperscript{***} & {\cellcolor{white}} \color{black} {\cellcolor[HTML]{3B4CC0}} \color[HTML]{F1F1F1} 0.69\textsuperscript{***} & {\cellcolor{white}} \color{black} {\cellcolor[HTML]{3B4CC0}} \color[HTML]{F1F1F1} 0.70\textsuperscript{***} & {\cellcolor{white}} \color{black} {\cellcolor[HTML]{3B4CC0}} \color[HTML]{F1F1F1} 0.78\textsuperscript{***} & {\cellcolor{white}} \color{black} {\cellcolor[HTML]{3B4CC0}} \color[HTML]{F1F1F1} 0.73\textsuperscript{***} & {\cellcolor{white}} \color{black} {\cellcolor[HTML]{3B4CC0}} \color[HTML]{F1F1F1} 0.73\textsuperscript{***} & {\cellcolor{white}} \color{black} {\cellcolor[HTML]{3B4CC0}} \color[HTML]{F1F1F1} 0.79\textsuperscript{***} & {\cellcolor{white}} \color{black} {\cellcolor[HTML]{3B4CC0}} \color[HTML]{F1F1F1} 0.75\textsuperscript{***} & {\cellcolor{white}} \color{black} {\cellcolor[HTML]{3B4CC0}} \color[HTML]{F1F1F1} 0.76\textsuperscript{***} \\
\bottomrule
\end{tabular}

}
\end{table}

\begin{table}[h]
\centering
\caption{Table of biases for fictional city names across all models. Significance markers reflect Bonferroni-corrected $\alpha$ levels for $m=183$ total tests (* $p_{\text{Bonf}} < 0.05$, ** $p_{\text{Bonf}} < 0.01$, *** $p_{\text{Bonf}} < 0.001$).}
\label{tab:A.3_bias_fracs_cities}
\resizebox{\textwidth}{!}{
\begin{tabular}{lllll||lll}
\toprule
 & \shortstack{Gemini \\ 2.5 Flash} & \shortstack{Gemini \\ 3 Flash} & \shortstack{Claude \\ Haiku 4.5} & \shortstack{GPT-5 \\ Mini} & \shortstack{Random \\ Pref Order 1} & \shortstack{Random \\ Pref Order 2} & \shortstack{Random \\ Pref Order 3} \\
Metric &  &  &  &  &  &  &  \\
\midrule
Starting Vowel (32\%) & {\cellcolor{white}} \color{black} {\cellcolor[HTML]{3B4CC0}} \color[HTML]{F1F1F1} 0.64 & {\cellcolor{white}} \color{black} {\cellcolor[HTML]{3B4CC0}} \color[HTML]{F1F1F1} 0.64 & {\cellcolor{white}} \color{black} {\cellcolor[HTML]{B2CCFB}} \color[HTML]{000000} 0.53 & {\cellcolor{white}} \color{black} {\cellcolor[HTML]{3B4CC0}} \color[HTML]{F1F1F1} 0.64 & {\cellcolor{white}} \color{black} {\cellcolor[HTML]{3B4CC0}} \color[HTML]{F1F1F1} 0.61 & {\cellcolor{white}} \color{black} {\cellcolor[HTML]{7699F6}} \color[HTML]{F1F1F1} 0.56 & {\cellcolor{white}} \color{black} {\cellcolor[HTML]{F7BA9F}} \color[HTML]{000000} 0.47 \\
Longer Length & {\cellcolor{white}} \color{black} {\cellcolor[HTML]{E8765C}} \color[HTML]{F1F1F1} 0.43 & {\cellcolor{white}} \color{black} {\cellcolor[HTML]{B40426}} \color[HTML]{F1F1F1} 0.39\textsuperscript{*} & {\cellcolor{white}} \color{black} {\cellcolor[HTML]{D0473D}} \color[HTML]{F1F1F1} 0.42 & {\cellcolor{white}} \color{black} {\cellcolor[HTML]{C0282F}} \color[HTML]{F1F1F1} 0.41\textsuperscript{*} & {\cellcolor{white}} \color{black} {\cellcolor[HTML]{AFCAFC}} \color[HTML]{000000} 0.53 & {\cellcolor{white}} \color{black} {\cellcolor[HTML]{CF453C}} \color[HTML]{F1F1F1} 0.42\textsuperscript{*} & {\cellcolor{white}} \color{black} {\cellcolor[HTML]{C1D4F4}} \color[HTML]{000000} 0.52 \\
\bottomrule
\end{tabular}

}
\end{table}

\subsection{Regression Coefficients for Other Datasets} \label{app_sec:regression_coeffs_others}
In this section, we present the logistic regression coefficients for the Numbers and Cities datasets, complementing the results shown for the Names dataset in Figure~\ref{fig:B_log_reg} of the main text.

\begin{figure*}[htbp]
    \centering
    \includegraphics[width=0.95\linewidth]{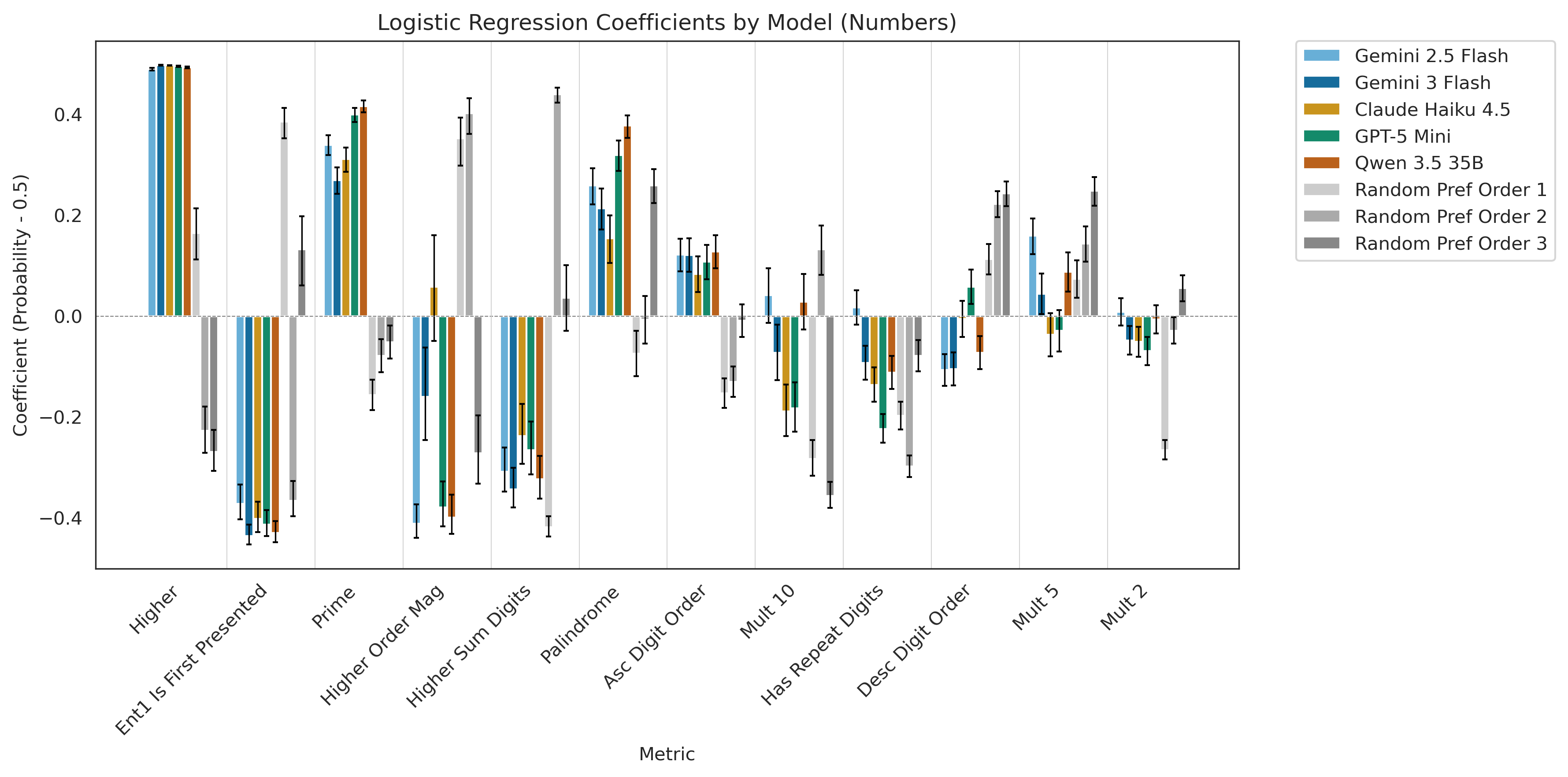}
    \caption{Logistic regression coefficients (with 95\% confidence intervals) for the Numbers dataset.}
    \label{app_fig:fig_B_numbers}
\end{figure*}

\begin{figure*}[htbp]
    \centering
    \includegraphics[width=0.95\linewidth]{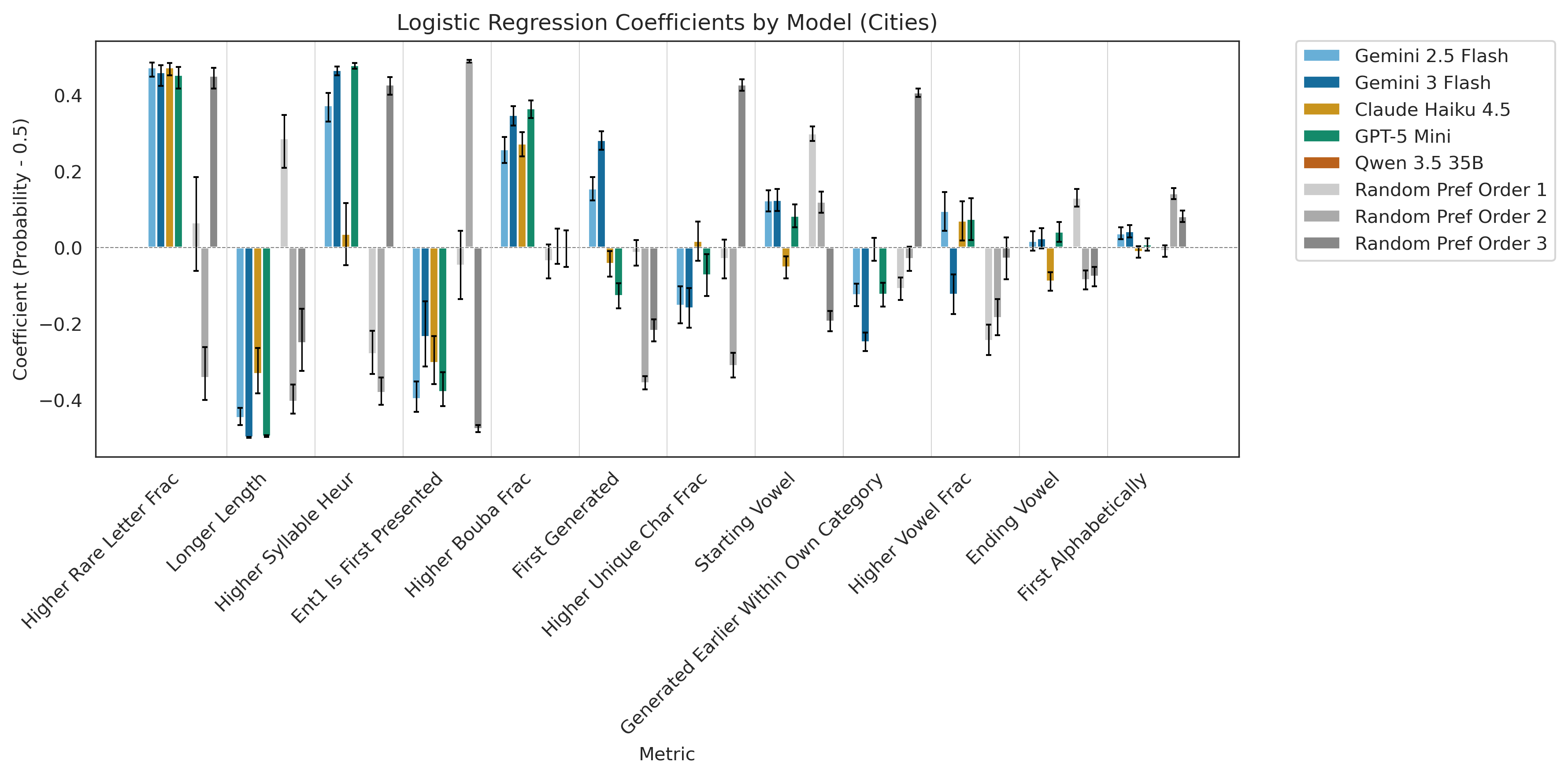}
    \caption{Logistic regression coefficients (with 95\% confidence intervals) for the Cities dataset.}
    \label{app_fig:fig_B_cities}
\end{figure*}

\subsection{Prompt Variation Histograms} \label{app_sec:prompt_hist_others}
These figures show the distribution of per-prompt winning fractions across all three datasets for Gemini 3 Flash, GPT-5 Mini, and Qwen 3.5 35B, respectively (representing the equivalent of Figure~\ref{fig:D_prompt_bias_hist} in the main text).

\begin{figure*}[htbp]
    \centering
    \includegraphics[width=0.95\linewidth]{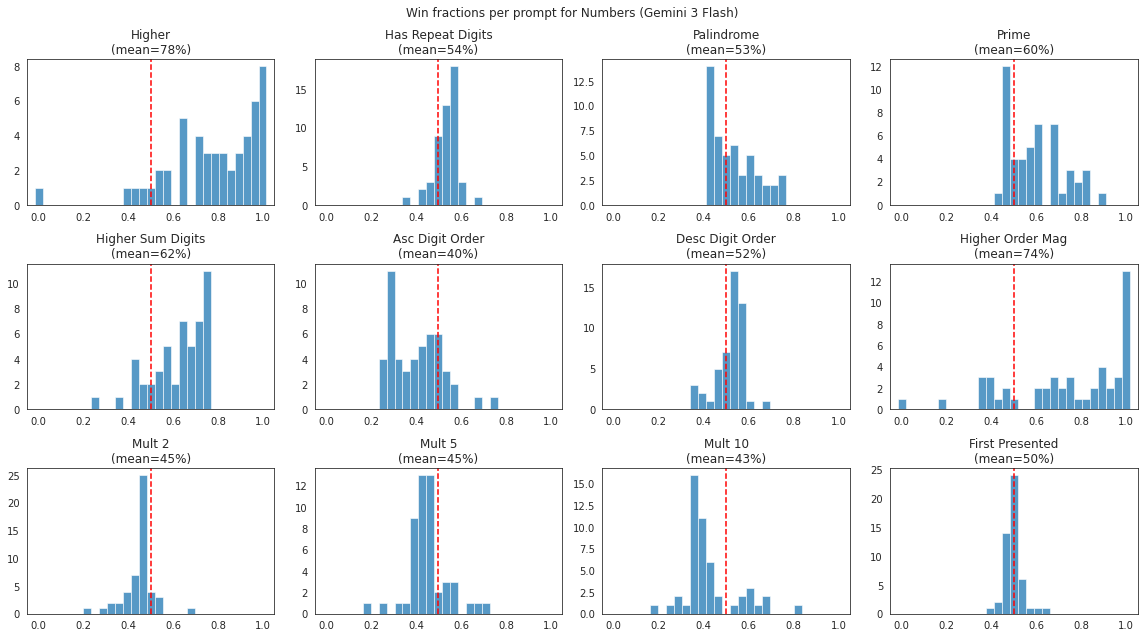}
    \caption{Distribution of induced preferences per prompt for Gemini 3 Flash.}
    \label{app_fig:fig_D_gdm}
\end{figure*}

\begin{figure*}[htbp]
    \centering
    \includegraphics[width=0.95\linewidth]{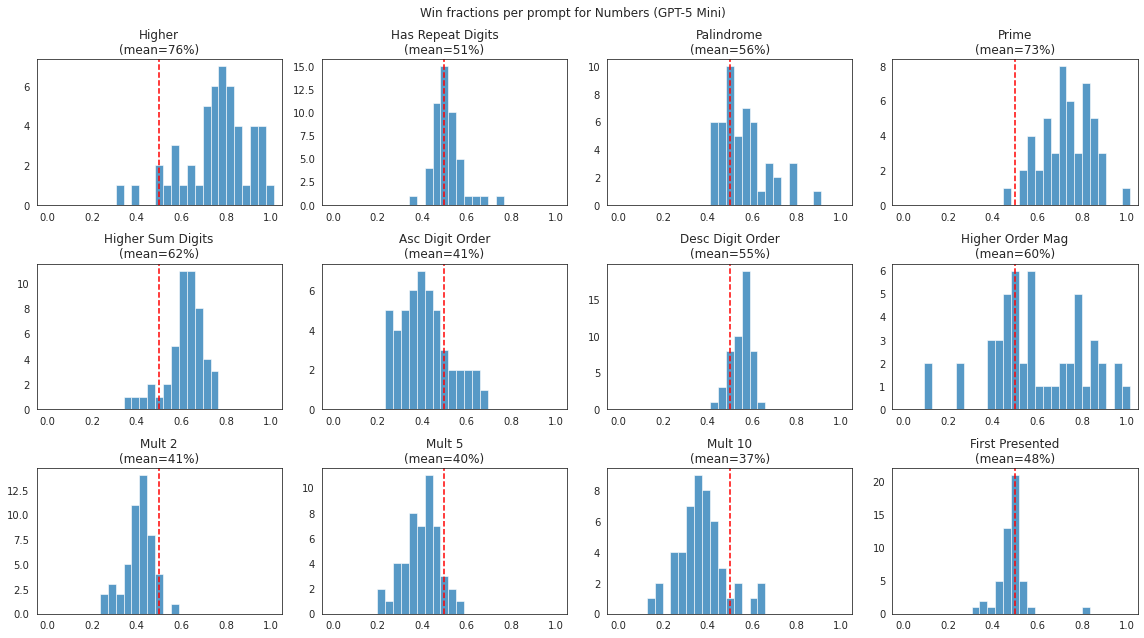}
    \caption{Distribution of induced preferences per prompt for GPT-5 Mini.}
    \label{app_fig:fig_D_openai}
\end{figure*}

\begin{figure*}[htbp]
    \centering
    \includegraphics[width=0.95\linewidth]{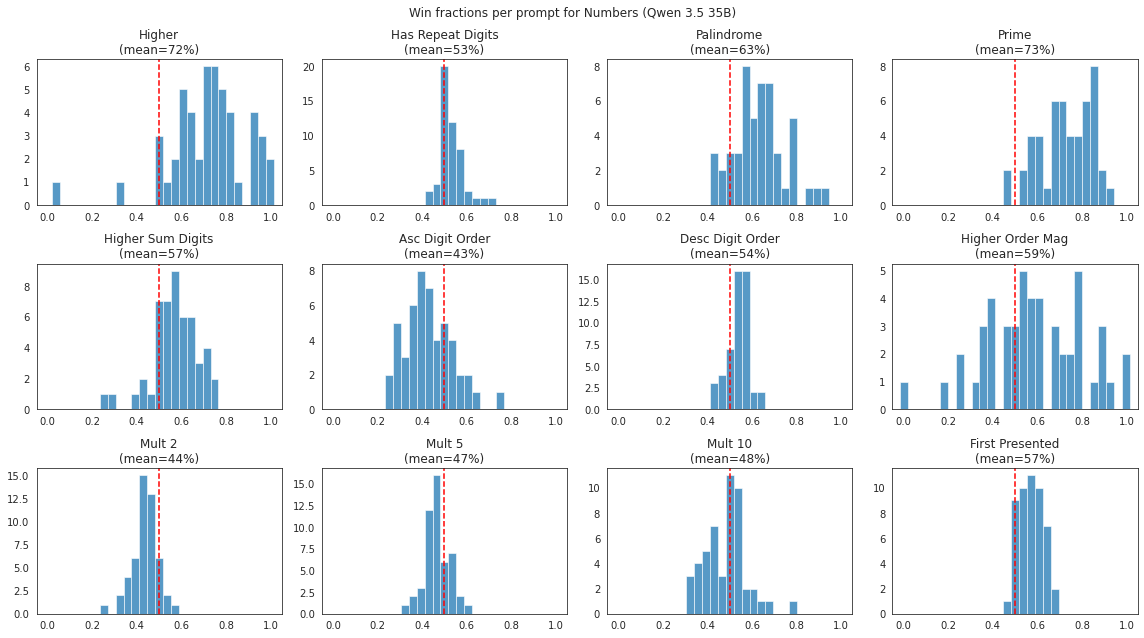}
    \caption{Distribution of induced preferences per prompt for Qwen 3.5 35B.}
    \label{app_fig:fig_D_qwen}
\end{figure*}

\newpage
\subsection{Prompt Robustness Across Datasets} \label{app_sec:prompt_robustness_others}
\begin{figure*}[htbp]
    \centering
    \begin{subfigure}{0.48\linewidth}
      \includegraphics[width=1\linewidth]{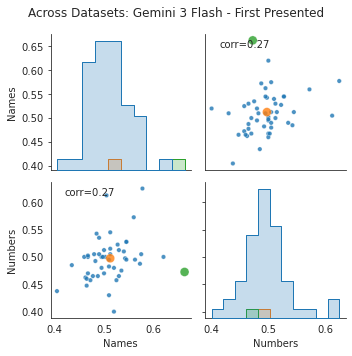}
      \caption{First Presented}
    \end{subfigure}
    \hfill
    \begin{subfigure}{0.48\linewidth}
      \includegraphics[width=1\linewidth]{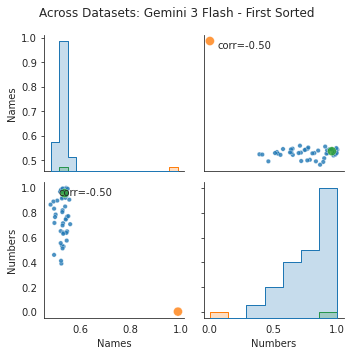}
      \caption{First Sorted}
    \end{subfigure}
    \caption{Correlation of prompt preference between datasets for Gemini 3 Flash.}
    \label{app_fig:fig_E_gdm}
\end{figure*}

\begin{figure*}[htbp]
    \centering
    \begin{subfigure}{0.48\linewidth}
      \includegraphics[width=1\linewidth]{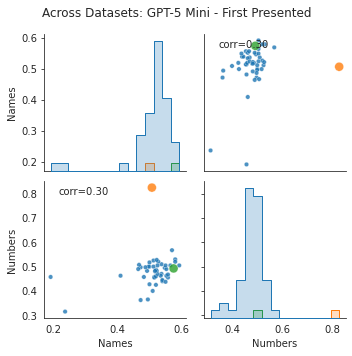}
      \caption{First Presented}
    \end{subfigure}
    \hfill
    \begin{subfigure}{0.48\linewidth}
      \includegraphics[width=1\linewidth]{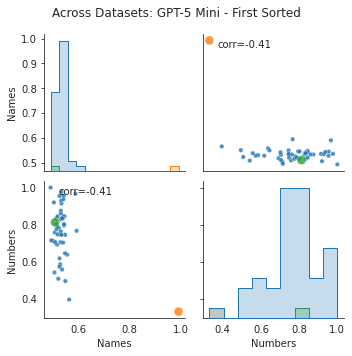}
      \caption{First Sorted}
    \end{subfigure}
    \caption{Correlation of prompt preference between datasets for GPT-5 Mini.}
    \label{app_fig:fig_E_openai}
\end{figure*}

Figures~\ref{app_fig:fig_E_gdm} and \ref{app_fig:fig_E_openai} show the correlation of prompt-dependent preferences across different datasets for Gemini 3 Flash and GPT-5 Mini (representing the equivalent of Figure\ref{fig:E_robust} in the main text).

\newpage
\subsection{Prompt Behavior Correlations Across Models} \label{app_sec:prompt_corr_models}
\begin{figure*}[htbp]
    \centering
    \begin{subfigure}{0.76\linewidth}
      \includegraphics[width=1\linewidth]{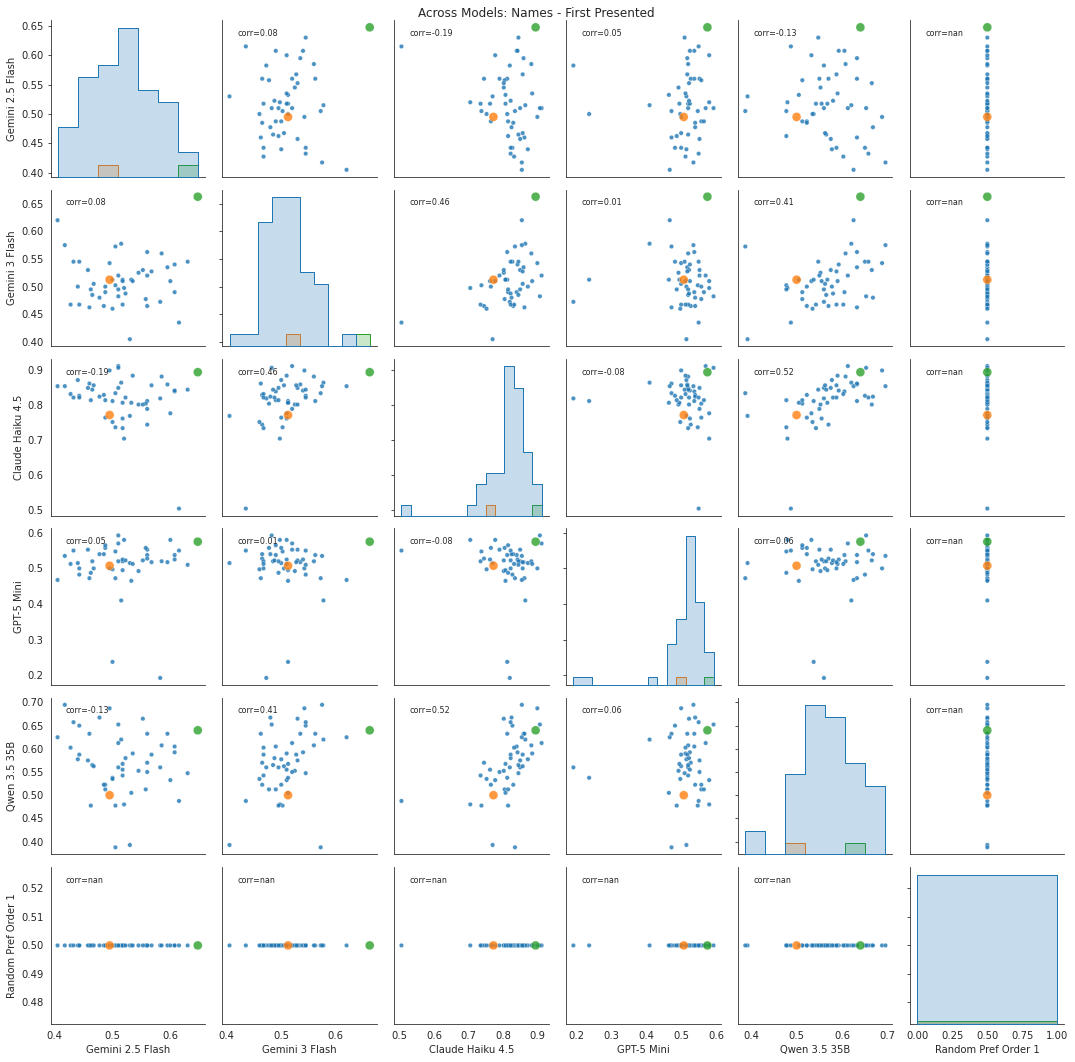}
      \caption{First Presented}
    \end{subfigure}
    \hfill
    \begin{subfigure}{0.76\linewidth}
      \includegraphics[width=1\linewidth]{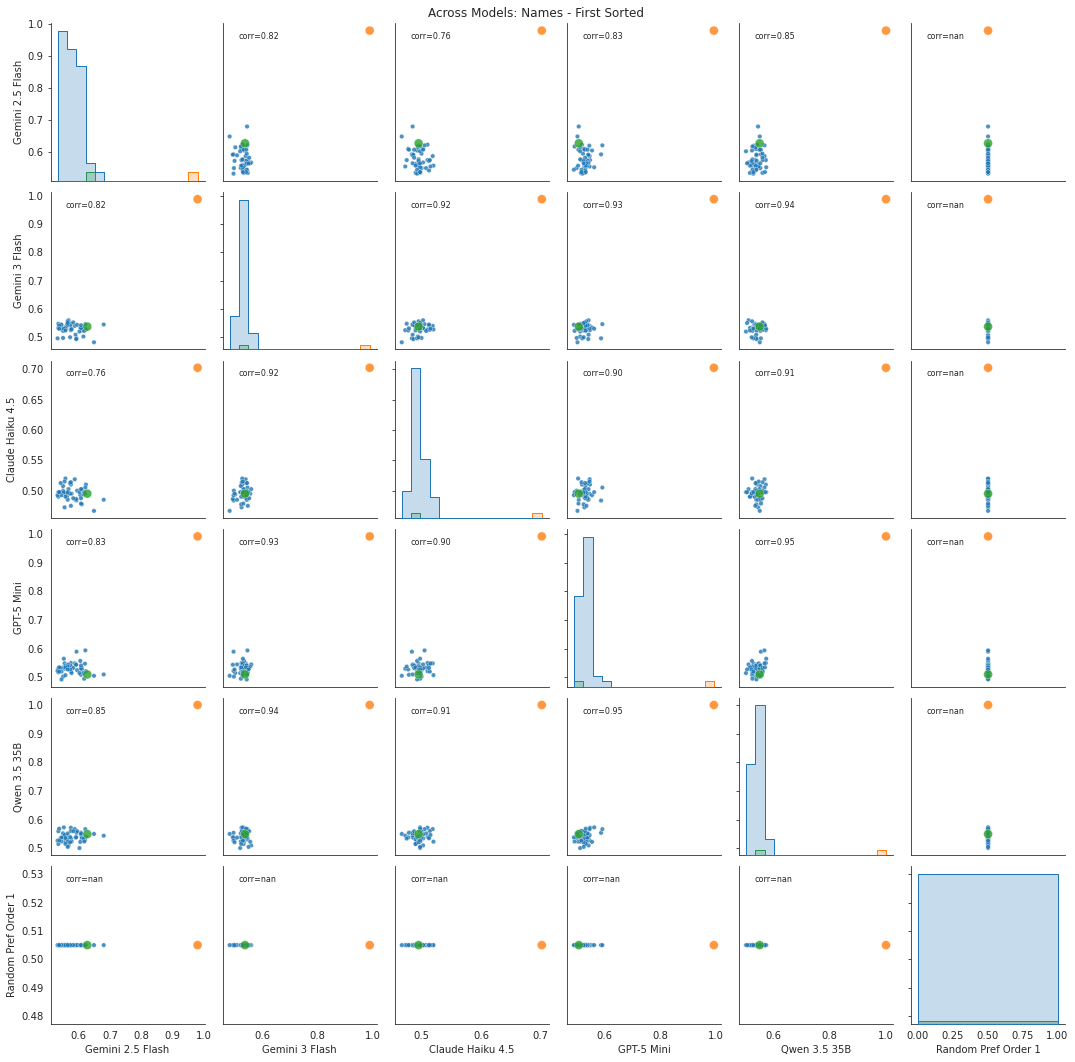}
      \caption{First Sorted}
    \end{subfigure}
    \caption{Prompt behavior correlations across all models for the Names dataset.}
    \label{app_fig:fig_E_models_names}
\end{figure*}

\begin{figure*}[htbp]
    \centering
    \begin{subfigure}{0.76\linewidth}
      \includegraphics[width=1\linewidth]{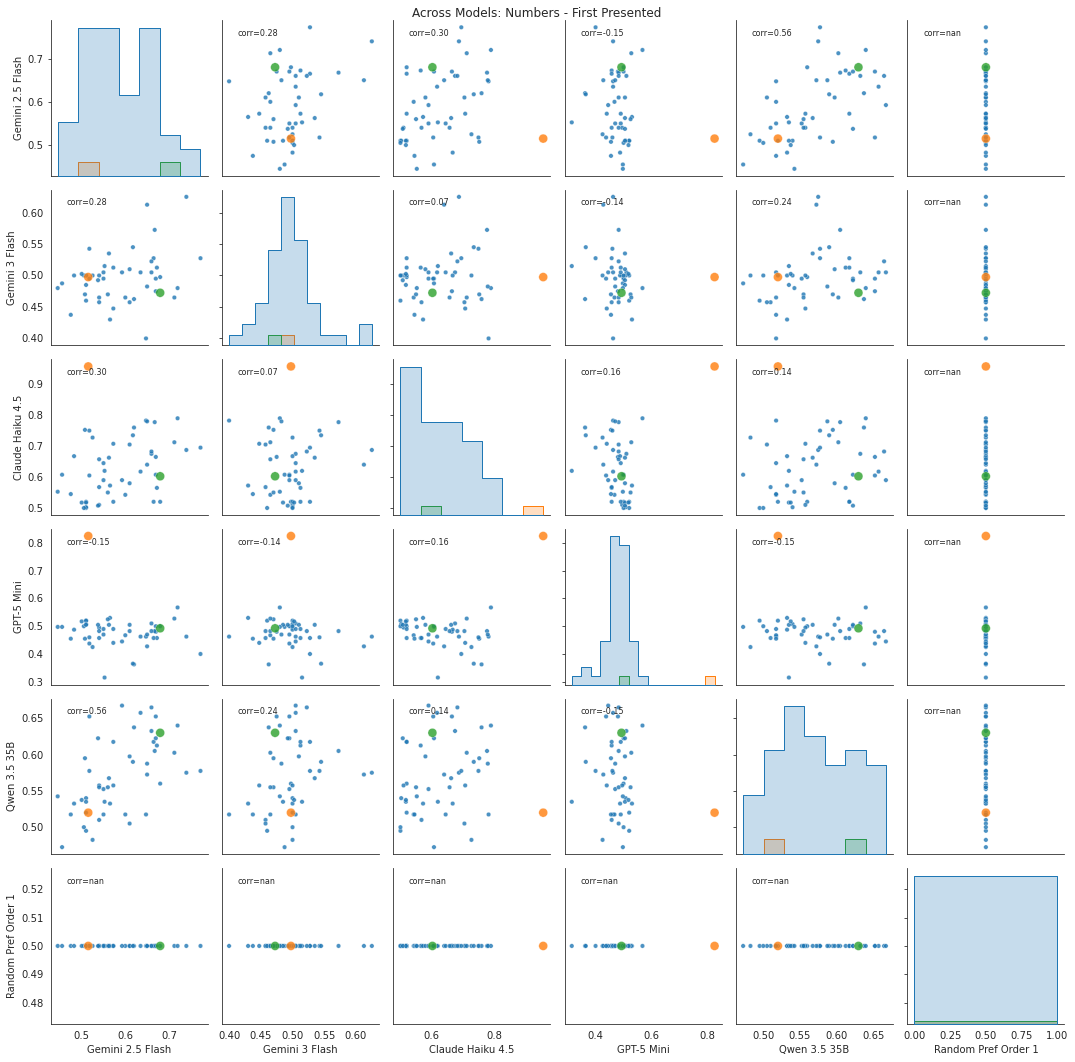}
      \caption{First Presented}
    \end{subfigure}
    \hfill
    \begin{subfigure}{0.76\linewidth}
      \includegraphics[width=1\linewidth]{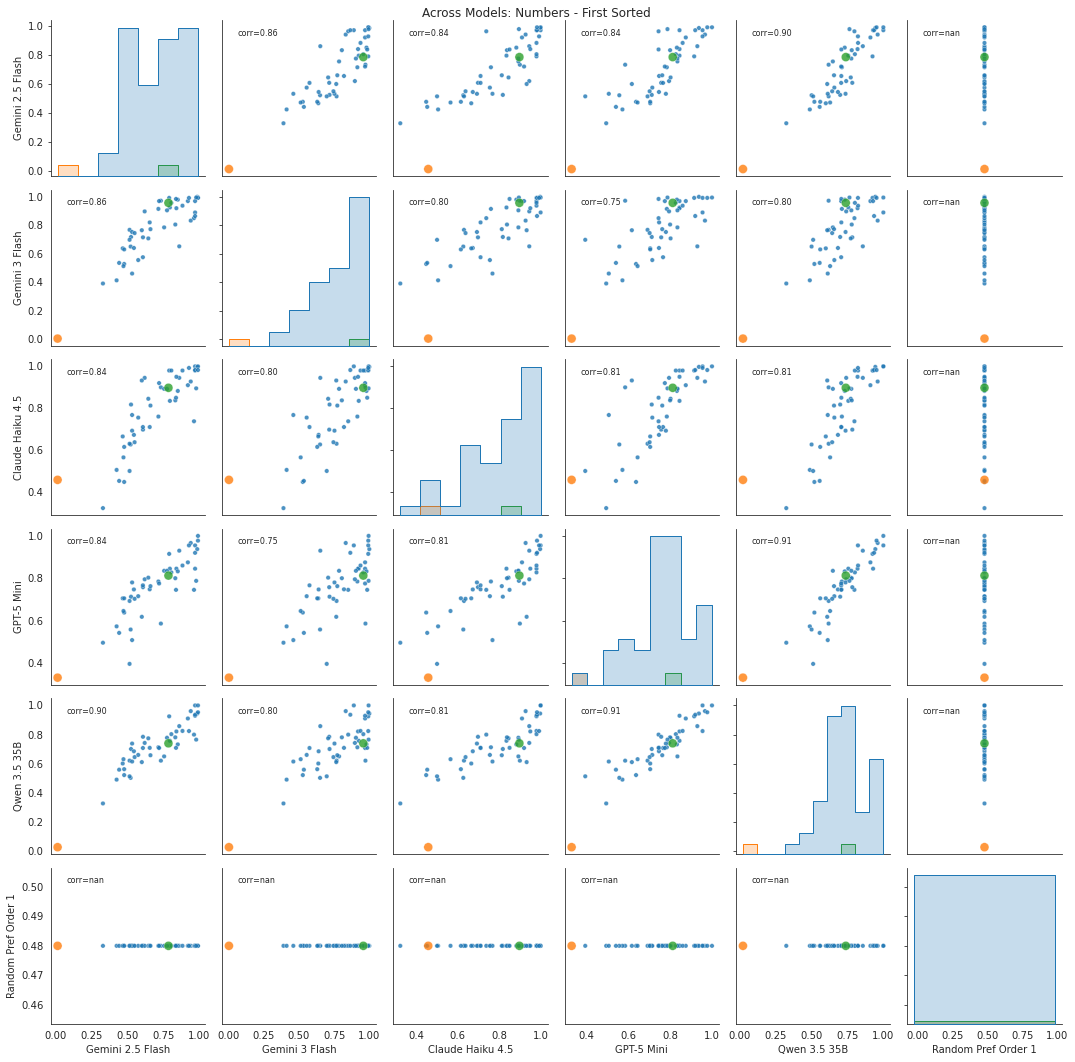}
      \caption{First Sorted}
    \end{subfigure}
    \caption{Prompt behavior correlations across all models for the Numbers dataset.}
    \label{app_fig:fig_E_models_numbers}
\end{figure*}

Finally, we examine how prompt-dependent preferences correlate *between* different model families for the same dataset. Figures~\ref{app_fig:fig_E_models_names} and \ref{app_fig:fig_E_models_numbers} show the PairGrid correlations across all evaluated SOTA models for the Names and Numbers datasets, respectively.


\clearpage
\section*{NeurIPS Paper Checklist}

\begin{enumerate}

\item {\bf Claims}
    \item[] Question: Do the main claims made in the abstract and introduction accurately reflect the paper's contributions and scope?
    \item[] Answer: \answerYes{}
    \item[] Justification: The abstract and Section~\ref{sec:intro} (Introduction) accurately describe the paper's scope—exploring subjective LLM biases—and its main contributions: the identification of dozens of biases (Section~\ref{sec:bias}), the impact of prompt variation (Section~\ref{sec:prompts}), and the interplay between ideation and pruning (Section~\ref{sec:ideation}).

\item {\bf Limitations}
    \item[] Question: Does the paper discuss the limitations of the work performed by the authors?
    \item[] Answer: \answerYes{}
    \item[] Justification: The paper has a limitations section in Section~\ref{sec:conclusion}. Additionally, throughout the text, the paper discusses scope limitations, such as focusing on stylized tasks (Section~\ref{sec:intro}) and short-form items (Section~\ref{sec:methods}) to isolate specific biases. 

\item {\bf Theory assumptions and proofs}
    \item[] Question: For each theoretical result, does the paper provide the full set of assumptions and a complete (and correct) proof?
    \item[] Answer: \answerNA{}
    \item[] Justification: The paper is an empirical study of LLM behavior in subjective decision-making tasks and does not contain any theoretical results, theorems, or proofs.

    \item {\bf Experimental result reproducibility}
    \item[] Question: Does the paper fully disclose all the information needed to reproduce the main experimental results of the paper to the extent that it affects the main claims and/or conclusions of the paper (regardless of whether the code and data are provided or not)?
    \item[] Answer: \answerYes{}
    \item[] Justification: The paper provides comprehensive details on the datasets, prompt expansion, and experimental framework in Section~\ref{sec:methods} and Appendix~\ref{app_sec:comp_object_tables} \& \ref{app_sec:prompts_used}. Appendix~\ref{app_sec:prompts_used} covers the exact prompts used in all phases.

\item {\bf Open access to data and code}
    \item[] Question: Does the paper provide open access to the data and code, with sufficient instructions to faithfully reproduce the main experimental results, as described in supplemental material?
    \item[] Answer: \answerNo{}
    \item[] Justification: We cannot release all of the code, but all critical experimental parameters, datasets, and prompts are reproduced in full in Appendix~\ref{app_sec:comp_object_tables} and Appendix~\ref{app_sec:prompt_tables}.

\item {\bf Experimental setting/details}
    \item[] Question: Does the paper specify all the training and test details (e.g., data splits, hyperparameters, how they were chosen, type of optimizer) necessary to understand the results?
    \item[] Answer: \answerYes{}
    \item[] Justification: We specify all experimental parameters including model versions, sampling methods, temperature settings ($T=1$, except for in Appendix~\ref{app:detailed-tables} where other temperatures are analyzed), and the pairwise comparison task in Section~\ref{sec:methods}.

\item {\bf Experiment statistical significance}
    \item[] Question: Does the paper report error bars suitably and correctly defined or other appropriate information about the statistical significance of the experiments?
    \item[] Answer: \answerYes{}
    \item[] Justification: We report statistical significance using Bonferroni-corrected one-sample t-tests against a 0.5 baseline and Levene's statistic, detailed in Section~\ref{sec:bias} and Section~\ref{sec:prompts}, using conservative sample size estimates. 

\item {\bf Experiments compute resources}
    \item[] Question: For each experiment, does the paper provide sufficient information on the computer resources (type of compute workers, memory, time of execution) needed to reproduce the experiments?
    \item[] Answer: \answerYes{}
    \item[] Justification: The experiments were conducted using cloud-based LLM APIs (Gemini, Claude, GPT) and open-source LLMs (Qwen). We specify the models and settings in Section~\ref{sec:methods} and Appendix~\ref{app_sec:prompts_used}, which defines the compute requirements.
    
\item {\bf Code of ethics}
    \item[] Question: Does the research conducted in the paper conform, in every respect, with the NeurIPS Code of Ethics \url{https://neurips.cc/public/EthicsGuidelines}?
    \item[] Answer: \answerYes{}
    \item[] Justification: The research conforms to the NeurIPS Code of Ethics as it is a standard study of LLM behavior and biases using synthetic and publicly available demographic data in Section~\ref{sec:methods}.

\item {\bf Broader impacts}
    \item[] Question: Does the paper discuss both potential positive societal impacts and negative societal impacts of the work performed?
    \item[] Answer: \answerYes{}
    \item[] Justification: The paper discusses potential positive and negative societal impacts of researching LLM subjective biases, including standardizing neutral assistance and the risk of prompt exploits, detailed in Appendix~\ref{app_sec:broader_impacts}.
    
\item {\bf Safeguards}
    \item[] Question: Does the paper describe safeguards that have been put in place for responsible release of data or models that have a high risk for misuse (e.g., pre-trained language models, image generators, or scraped datasets)?
    \item[] Answer: \answerNA{}
    \item[] Justification: The research does not involve high-risk data or the release of high-risk models, utilizing only standard public APIs and public census data.

\item {\bf Licenses for existing assets}
    \item[] Question: Are the creators or original owners of assets (e.g., code, data, models), used in the paper, properly credited and are the license and terms of use explicitly mentioned and properly respected?
    \item[] Answer: \answerYes{}
    \item[] Justification: Creators of existing assets are credited via citations, including US Census names, SSA data, and model configurations mapped in Appendix~\ref{app_sec:model_configs} (Table~\ref{tab:model_configs}). Licenses for public datasets are standard for US government work.

\item {\bf New assets}
    \item[] Question: Are new assets introduced in the paper well documented and is the documentation provided alongside the assets?
    \item[] Answer: \answerYes{}
    \item[] Justification: The paper introduces new assets—50 prompts and 200 fictional city names—which are reproduced in full in Appendix~\ref{app_sec:comp_object_tables} and Appendix~\ref{app_sec:prompt_tables}.

\item {\bf Crowdsourcing and research with human subjects}
    \item[] Question: For crowdsourcing experiments and research with human subjects, does the paper include the full text of instructions given to participants and screenshots, if applicable, as well as details about compensation (if any)? 
    \item[] Answer: \answerNA{}
    \item[] Justification: The research does not involve crowdsourcing or human subjects, relying entirely on LLM-driven experiments in Section~\ref{sec:methods}.

\item {\bf Institutional review board (IRB) approvals or equivalent for research with human subjects}
    \item[] Question: Does the paper describe potential risks incurred by study participants, whether such risks were disclosed to the subjects, and whether Institutional Review Board (IRB) approvals (or an equivalent approval/review based on the requirements of your country or institution) were obtained?
    \item[] Answer: \answerNA{}
    \item[] Justification: IRB approval is not applicable as the study does not involve research with human subjects.

\item {\bf Declaration of LLM usage}
    \item[] Question: Does the paper describe the usage of LLMs if it is an important, original, or non-standard component of the core methods in this research? Note that if the LLM is used only for writing, editing, or formatting purposes and does \emph{not} impact the core methodology, scientific rigor, or originality of the research, declaration is not required.
    \item[] Answer: \answerYes{}
    \item[] Justification: The paper explicitly describes the usage of LLMs for generating city names, expanding prompts, and categorizing city names and prompts in Section~\ref{sec:methods} and Section~\ref{sec:prompts}.

\end{enumerate}

\end{document}